\documentclass{article}
\usepackage{jfrExamplee}
\usepackage{graphicx}
\usepackage{apalike}
\usepackage{setspace}
\usepackage{xcolor}
\usepackage{todonotes} 
\usepackage{svg}
\usepackage{comment}
\usepackage{graphicx}
\usepackage{amsmath}
\usepackage{amssymb}
\usepackage{float}
\usepackage{etoolbox}
\usepackage{url}
\usepackage{mathrsfs}
\usepackage{algorithm}
\usepackage{algpseudocode}
\algnewcommand{\LineComment}[1]{\State \(\triangleright\) #1}
\usetikzlibrary{shapes.geometric, arrows, positioning}
\usepackage{cite}

\usepackage[nolist]{acronym}

\acrodef{MCTS}[MCTS]{Monte Carlo tree search}
\acrodef{RNN}[RNN]{recurrent neural network}
\acrodef{GRNN}[GRNN]{generative \acl{RNN}}
\acrodef{RL}[RL]{reinforcement learning}
\acrodef{SARL}[SARL]{socially attentive \acl{RL}}
\acrodef{WMR}[WMR]{wheeled mobile robot}
\acrodef{CNN}[CNN]{convolutional neural network}
\acrodef{LIDAR}[LIDAR]{light detection and ranging}\acused{LIDAR}
\acrodef{FS}[FS]{fail-safe}
\acrodef{RED}[RED]{\acl{RNN} encoder-decoder}
\acrodef{PRM}[PRM]{probabilistic roadmap}
\acrodef{RHEC}[RHEC]{receding horizon estimation and control}
\acrodef{PID}[PID]{proportional-integral-derivative control}\acused{PID}
\acrodef{PF}[PF]{potential field}
\acrodef{GNSS}[GNSS]{global navigation satellite system}\acused{GNSS}
\acrodef{IMU}[IMU]{inertial measurement unit}\acused{IMU}
\acrodef{ORCA}[ORCA]{optimal reciprocal collision avoidance}
\acrodef{FOV}[FOV]{field of view}

\title{Resource and Response Aware Path Planning for Long-term Autonomy of Ground Robots in Agriculture}

\author{Stuart Eiffert\thanks{Stuart Eiffert and Nathan Wallace contributed equally to this work; corresponding author: He Kong} , Nathan D. Wallace$^{*}$, He Kong, Navid Pirmarzdashti, and Salah Sukkarieh \\ 
Australian Centre for Field Robotics, University of Sydney \\
\texttt{\{s.eiffert,n.wallace,h.kong,navidp,salah\}@acfr.usyd.edu.au} \\
}

\begin{document}

\maketitle

\begin{abstract}

Achieving long-term autonomy for mobile robots operating in real-world unstructured environments such as farms remains a significant challenge. This is made increasingly complex in the presence of moving humans or livestock. These environments require a robot to be adaptive in its immediate plans, accounting for the state of nearby individuals and the response that they might have to the robot's actions. Additionally, in order to achieve longer-term goals, consideration of the limited on-board resources available to the robot is required, especially for extended missions such as weeding an agricultural field. To achieve efficient long-term autonomy, it is thus crucial to understand the impact that online dynamic updates to an energy efficient offline plan might have on resource usage whilst navigating through crowds or herds. To address these challenges, a hierarchical planning framework is proposed, integrating an online local dynamic path planner with an offline longer-term objective-based planner. This framework acts to achieve long-term autonomy through awareness of both dynamic responses of individuals to a robot's motion and the limited resources available. This paper details the hierarchical approach and its integration on a robotic platform, including a comprehensive description of the planning framework and associated perception modules. The approach is evaluated in real-world trials on farms, requiring both consideration of limited battery capacity and the presence of nearby moving individuals. These trials additionally demonstrate the ability of the framework to adapt resource use through variation of the local dynamic planner, allowing adaptive behaviour in changing environments. A summary video is available at https://youtu.be/DGVTrYwJ304.

\end{abstract}

\section{Introduction}

Mobile robots are increasingly being deployed in unstructured environments where they are required to operate over extended periods of time. These applications often involve continuously performing routine tasks such as cleaning indoor environments, warehouse package delivery, or weeding of agricultural fields. The presence of moving individuals, such as pedestrians or livestock, can make these applications increasingly challenging. Mobile robots need to be aware of the likely response of any nearby individuals both for safety and efficiency as they navigate.
This is especially important when operating in crowds or herds where non response-aware planning approaches can result in the `frozen robot' problem \cite{Trautman2010}.
Consideration of the limited resources available to a mobile robot---including time and energy---is also important during extended operation to ensure completion in a timely manner, and avoid premature exhaustion of any mission critical resources. However, optimisation may require a trade off between conflicting resources, especially in dynamic environments. A more time efficient path through a crowd may require the robot to travel greater distances, and so use more energy.
In order to achieve efficient long-term autonomy in real-world environments, a planning framework that is aware of both the limited resources available to a robot and the response of nearby moving individuals, is required.

Of particular interest in this paper are applications of mobile ground robots in agriculture and large-scale farming. These applications demand completion of a large variety of essential tasks such as soil sampling, weeding, crop observation, and recharging, which are often dispersed widely over large geographic areas. Additionally, these agricultural tasks are frequently carried out in the presence of humans and livestock. In these applications, consideration of resources available to a robot is required both during the formation of long-term mission plans, and in the computation of dynamic updates to the plan. Operating in unstructured environments often requires the balancing of on-board energy against total mission time. A typical example of this is planning through undulating terrain, where the fastest route is not always the most energy-efficient. Similarly, navigating through a crowd of agents---such as a herd of livestock---requires an understanding of how online deviations from an energy optimal reference path can impact the robot's resource usage; a crucial consideration during the completion of time-critical tasks such as harvesting.



In this work, we propose a hierarchical path planning framework to enable the long-term autonomy of mobile ground robots in unstructured and dynamic environments, subject to resource constraints of energy and time. We build upon the contributions of our prior work \cite{Eiffert2020b} and \cite{Eiffert2020c}, extending the description of the overall framework as well as detailing the perception modules used for object detection, tracking, and static mapping. This framework uses a resource-aware long-term planner for the formation of strategic-level plans to allow for navigation between goal locations subject to energy constraints. An online response-aware local dynamic planner is utilised alongside the offline long-term planner. This enables the updating of the strategic plan both in response to unforeseen static obstacles, and with consideration of the response of detected nearby moving individuals to the robot's motion. These planners are used in combination with a higher-level mode switching module, allowing adaptation of the robot's behaviour dependent on the detection of nearby agents and obstacles. 

In addition, we describe a novel perception pipeline used for the simultaneous 3D object tracking and static mapping required by the local dynamic planner, and detail the integration of this pipeline into the hierarchical planning framework. 
Whilst previous work has demonstrated the simultaneous detection and mapping of static and dynamic elements in unstructured environment, the extension of these perception pipelines for use within a dynamic planning framework for long term autonomy has not yet been demonstrated.



The performance of the proposed approach is evaluated in a series of simulated and real world trials. Simulated testing is conducted in a high-fidelity dynamic simulation environment and has included comparison of performance with varying local dynamic planners. These tests highlight the ability of our planning framework to adapt its resource usage to changing constraints. A comparison of four different dynamic planning strategies---a response-aware planner, which uses \ac{MCTS} with \ac{GRNN} models of agent response \cite{Eiffert2020a}; \ac{SARL} \cite{chen2019crowd}, a state of the art dynamic planner; a traditional \ac{PF} planner; and a purely reactive \ac{FS} planner---outlines how each version can be used to optimise for either time or energy efficiency in varying crowd densities. The analysis in this work builds upon the results of our prior work \cite{Eiffert2020b}, indicating the potential of our proposed approach to allow adaptive robot behaviour dependent on both changing resource constraints and the presence of moving individuals during extended operation.

Real world demonstration and validation of the proposed approach's performance is conducted on the University of Sydney's Swagbot robotic platform; a robot designed for use in extended agricultural tasks, including the weeding of pastures alongside moving individuals. Evaluation includes the continuous navigation between sets of long-term goals throughout an unstructured agricultural field---in this scenario, representative of the locations of weeds to be sprayed using Swagbot's actuated weed sprayer. The robot was required to plan in an energy-efficient manner whilst in the presence of moving individuals and unknown obstacles. Further empirical evaluation in more densely populated environments is also presented, featuring repeated interactions in a pedestrian crowd. The real-world performance in terms of both safety and resource efficiency has been compared to the simulated results from our prior work \cite{Eiffert2020b}. We additionally perform evaluation of the perception module, discussing how an understanding of agent observation likelihood across the robot's observation space should be taken into consideration during dynamic path planning. A video summarising all trials is available at https://youtu.be/DGVTrYwJ304.

The results confirm that the proposed hierarchical planning framework is able to allow long-term autonomy of a mobile robot in unstructured environments.  Through a combination of resource and response-aware path planning, the safe and efficient navigation of dynamic environments with consideration of resource constraints has been achieved. 
The main contributions of this paper include:
\begin{enumerate}
    \item Comprehensive evaluation of our proposed planning framework, based on the sum of prior work, in both simulated and real-world trials to demonstrate safety around moving individuals and adaptive resource usage for extended autonomy in unstructured and dynamic agricultural environments.
    \item Simulated testing of the framework using varied local dynamic planners, including our prior MCTS-GRNN and a state of the art \ac{SARL} planner, to demonstrate framework adaptability and compare resource usage and tradeoffs.
    \item Additional real-world testing of the framework in order to evaluate behaviour during continuous interactions with moving individuals.
    \item Detailed evaluation of a perception module for simultaneous 3D object tracking and static mapping of an unstructured agricultural environment in real world trials.
\end{enumerate}

The remainder of the paper is structured as follows: Section \ref{section:background} describes prior work in dynamic path planning and resource constrained path planning, focusing on agricultural applications; Section \ref{section:approach} describes the proposed hierarchical planning framework, detailing each module; Section \ref{section:experiments} presents the experimental platform and testing methodology; Section \ref{section:results} summarises experimental results with discussion in Section \ref{section:discussion}; and Section \ref{section:conclusion} presents the conclusions.








\section{Background}
\label{section:background}


The application of mobile robots in real-world environments has seen significant growth in recent years. This is particularly true in more structured environments such as  on-road autonomous vehicles \cite{Litman2020} and indoor service robots in shared pedestrian environments such as retail stores and warehouses\cite{Inam2018}. However, the same levels of adoption have not yet been achieved in less structured environments, such as those seen in agriculture. Here, mobile robotic operations have largely been limited to tasks based in crop rows \cite{Bechar2017} or orchards \cite{Carpio2020}, where dynamic elements such as moving individuals are not a critical consideration, and the environment structure lends itself to repetitive motion patterns. Similar work has applied `teach and repeat' approach for long term adaptive route following in unstructured environments in the presence of moving agents and \cite{icpPlanner}. However, this work is unable to differentiate between static and dynamic obstacles, treating all moving individuals as static obstacles and updating an obstacle map each planning step. Additionally, these works have not yet addressed the requirement of resource awareness during planning, necessary for efficient long term autonomy. Whilst the ability to detect obstacles and agents in these unstructured and uncontrolled environments is improving \cite{Kragh2020}, perception remains a more challenging problem than in the domains of indoor or road-based use, where planar ground assumptions and lighting invariance help enable object and traversability detection \cite{krusi2017}.


Furthermore, the challenge of operating under limited resource budgets becomes much more significant when operating in off-road environments, such as large scale farms, where paths may not be specified and operating might often involve traversing non-uniform terrain. The average size of a farm in Australia in 2016 was \linebreak 4,331 ha \cite{ABS2016}. As such, long-term autonomy in these environments may require a mobile robot to travel vast distances between mission waypoints, requiring the management of conflicting time and energy usage constraints.

\subsection{Perception in Unstructured Environments}
\label{section:background_perception}
Safe and efficient navigation around moving individuals requires both an accurate estimate of the current state of the dynamic environment, and the ability to predict the future motion of any individuals, in order to plan accordingly. To achieve this in unstructured environments, it is necessary to both differentiate between traversable ground and obstacles, as well as between static and dynamic elements of the unknown environment.

In more structured environments, such as road based applications and indoor usage, direct 3D object detection in point clouds has made significant advancements in recent years \cite{qi2017pointnetplusplus, voxelnet}. State of the art learning-based techniques have been able to take advantage of large scale datasets of labelled 3D scenes specific to structured applications, such as the nuScenes \cite{nuscenes2019} and KITTI \cite{kitti} datasets. Multi-modal 3D object detectors such as frustum pointnets \cite{qi2017frustum} have combined these approaches with more mature 2D object detectors, however, these still require access to large datasets to train the 3D box segmentation and regression networks. These techniques are not directly applicable to agricultural applications as there do not currently exist any large scale labelled 3D datasets of all relevant objects, such as a livestock. Additionally, these existing state of the art 3D object detectors are generally not directly suitable for unstructured environments where segmentation of traversable ground and static obstacles is a non-trivial task, since in most road-based applications the flat ground assumption can be safely made over local areas. 

Previous work has demonstrated how ground segmentation in unstructured environments can be achieved through the use of piecewise planar surface fitting methods \cite{Asvadi2016} or conditional random fields \cite{Rummelhard2017}. More recent work has applied conditional random fields and 2D semantic segmentation to multi-modal sensor input for simultaneous ground plane segmentation and classwise 3D object detection \cite{Kragh2020}. The fusion of 2D visual input with point cloud data in this approach allows for the identification of traversable vegetation that may otherwise be detected as static obstacles when relying only on laser input. This is especially important in agricultural applications that require operating around weeds or long grass. These approaches to ground segmentation tend to incur significant computational costs, which is an important consideration for mobile robots operating under resource constrained extended missions. Additionally, the extension of these perception pipelines for use within a dynamic planning framework---one that accounts for the responses of these individuals to the robot's motion---has not yet been demonstrated in unstructured environments.

\subsection{Path Planning in Dynamic Environments}
\label{section:planning}

Whilst crowd motion prediction models based on hand-crafted features such as \acp{PF} or the social force model \cite{sfm} have been shown to achieve satisfactory performance when used for dynamic path planning, these methods can suffer from the `frozen robot' problem when in more complex environments \cite{Trautman2010}. `Response-aware' predictive models can overcome this problem by modelling the interactions between agents using methods such as reciprocal velocity obstacles \cite{VanDenBerg2011} or interacting Gaussian processes \cite{Trautman2010}. However, in more recent literature \cite{ivanovic2019, Eiffert2019} it has been shown that for the prediction of crowd motion during human-robot interactions, these methods can be outperformed by prediction models that use \acp{RNN} to learn the response of individuals to a robot's planned path from observed data.
\newpage
Use of \ac{RNN}-based models within sampling-based path planning systems \cite{Eiffert2020a} or model-based deep \ac{RL} methods \cite{chen2019crowd, Chen2020, Fan2020} has been proposed for dynamic path planners. Whilst these methods have been shown to enable robust dynamic path planning in unstructured and complex environments such as pedestrian crowds, they are yet to be extended for use  within a complete planning framework for long-term autonomy. Existing approaches to extended path planning in unstructured environments, such as in agricultural applications, have tended to not differentiate between static and dynamic elements of the environment when planning \cite{Carpio2020, Santos2020, icpPlanner}. By treating unforeseen static obstacles and moving individuals similarly during planning, these approaches are unable to model the response of other agents to a robot's future motion and are again subject to the `frozen robot' problem. The ability to continue effectively navigating when dynamic environments grow in complexity is critical to enable completion of missions in an efficient and timely manner. This is an especially important consideration when operating under resource constraints, such as will occur during extended missions.







\subsection{Resource Constrained Path Planning}
It is necessary to anticipate the cost of performing tasks and actions in the environment in order to best utilise mobile robots in the field. This is especially true for electric-powered \acp{WMR} in off-road and large scale environments, where the energy usage of the platform determines the robot's range and maximum operational time. This problem has been the topic of our previous work on modelling the energy cost of \ac{WMR} motion \cite{Wallace_WROCO}.

The challenge of path planning with consideration of energy usage has seen significant attention in recent research, with techniques developed that utilise cost models for energy use in point-to-point and coverage path planning \cite{Qian2010, Tokekar2014, Mei2006, Isler2018, Sun2020}, more efficient tracking of human-piloted reference paths \cite{Gao2020}, and planning of multi-stage paths in uneven off-road environments \cite{Wallace_Agricontrol}. Extending these works for long-term autonomy under resource constraints has previously been modelled as a variant of the orienteering problem in applications such as persistent environmental monitoring \cite{Yu2016} or data collection \cite{Faigl2017}. Similar work has explored approaches allowing resource recharging in transportation networks \cite{Cassandras2018} and logistics \cite{Laporte2013}. Our recent work on the orienteering problem with replenishment (OPR) \cite{Wallace_CASE2020} provides a generalised approach intended to handle revisiting an arbitrary number of recharging stations, while optimising for the total completion time of each task. This has been motivated by its application to time critical tasks in agriculture, such as harvesting  which often requires completion of entire fields within several hours to fit within logistic constraints, or the regular herding of livestock between fields at specified times.
The remainder of this work presents an approach to long term autonomy under resource limitations in a dynamic environment, analysing the impact and trade-offs of resource usage between varying dynamic path planning approaches. Recharging has not yet been included, which will require active perception docking and is left to future work.

\section{System Framework}
\label{section:approach}
The proposed hierarchical planning framework combines a local online dynamic planning module with a long-term offline planner, in order to allow extended autonomy in unstructured and dynamic environments. The work expands our prior work \cite{Eiffert2020b}, extending the description of the overall framework, as well as detailing the perception modules for object detection, tracking, and static mapping; utilised both by the local planner and by a \ac{FS} collision avoidance module. 
 \begin{figure}[t]
    \centering
	\includegraphics[width=16.5cm,height=6.6cm]{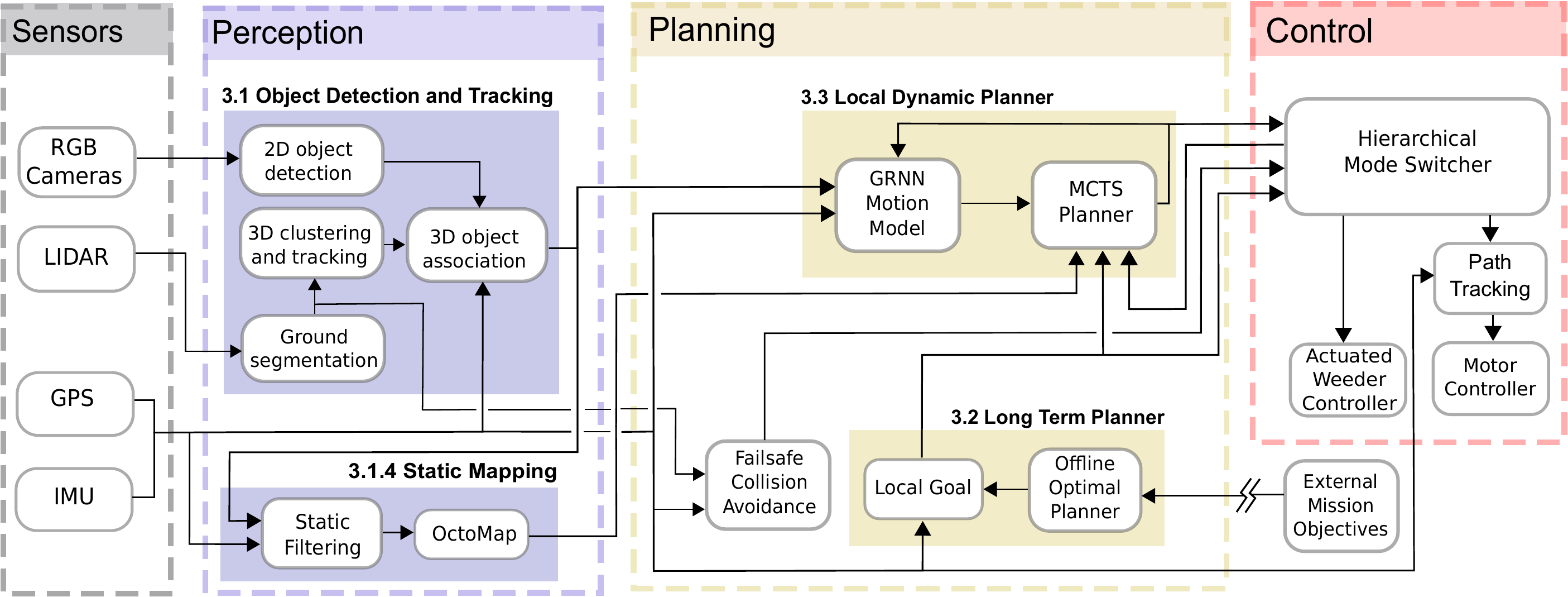}
	\setlength{\belowcaptionskip}{-5pt}
	\caption{\textit{System overview of the hierarchical framework, illustrating the communication between the hierarchical mode controller (red)and each planning module (yellow), including the long-term resource-aware planner, Local Dynamic response-aware planner and the \ac{FS} collision module. Mission objectives are provided externally to the long-term planner. The local dynamic planner module is shown here implementing our prior \ac{MCTS}-\ac{GRNN} planner \cite{Eiffert2020a}.}}
	\label{system_overview}
\end{figure}
Fig. \ref{system_overview} illustrates how each module is used and communicates within the hierarchical framework. A hierarchical mode switcher takes input from the long-term planner and dynamic planner, and determines which reference path to pass to the path tracking module, which directly controls the robot's motion. This is decided based on proximity to detected agents and to the local goal. 
When no dynamic agents or obstacles are detected within an 8 m radius of the robot, the local goal from the long-term planner is used directly as reference. Otherwise, the output of the dynamic planner, which tracks the local goal whilst avoiding dynamic agents and obstacles, is used instead.  
The hierarchical mode switcher also takes input from the \ac{FS} collision avoidance module, which stops all robot motion when an agent or obstacle is detected within a 2 m radius of the robot. Mission objective waypoints and information regarding the operating environment, such as terrain data and no-go areas, are provided to the long-term planner from an external source.


\subsection{Perception Pipeline}
\label{section:perception}
\begin{figure}[t]
    \centering
	\includegraphics[width=16.5cm,height=7.4cm]{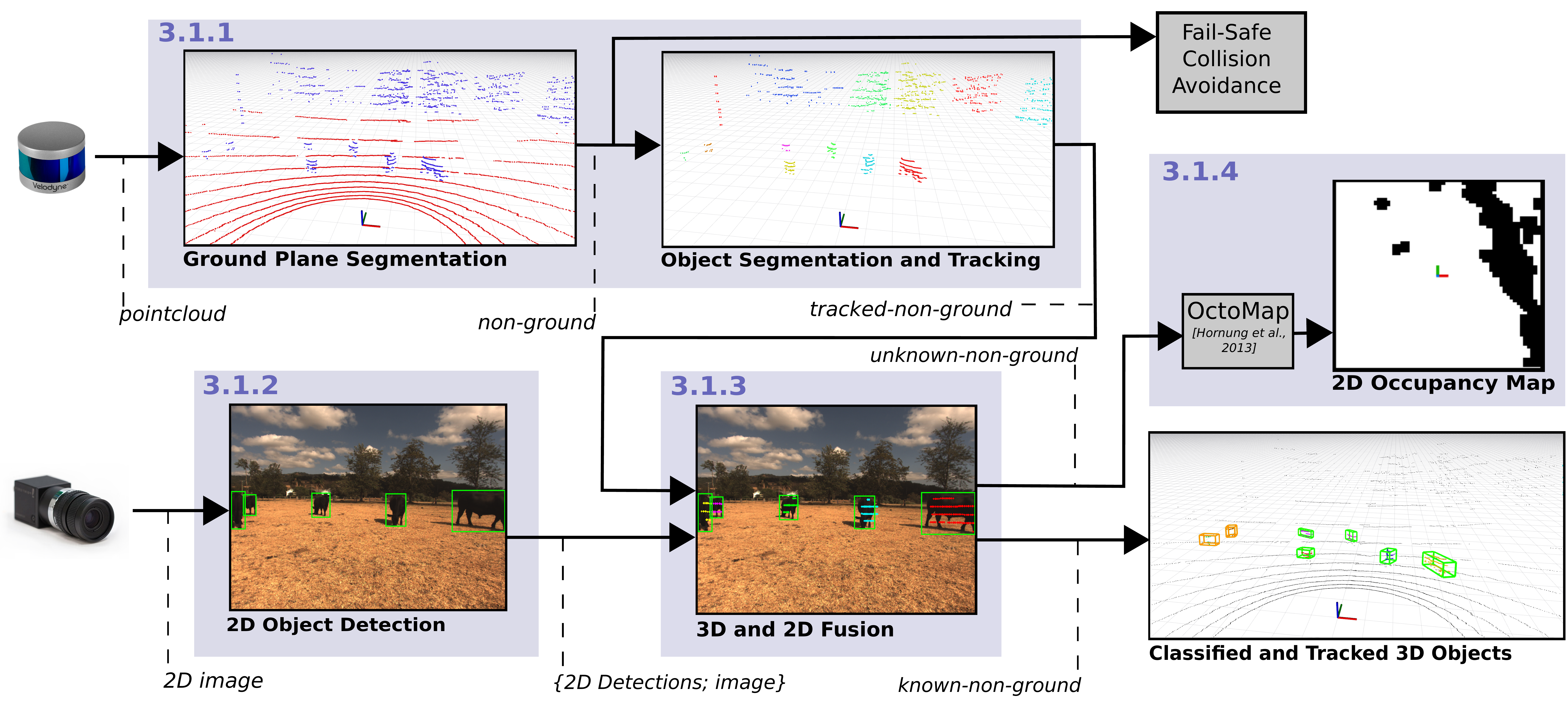}
	\setlength{\belowcaptionskip}{0pt}
	\caption{\textit{Perception pipeline used for 3D object detection and tracking. 2D object detection is performed synchronously with 3D point cloud ground plane segmentation, object segmentation, and tracking. The tracked 3D objects are projected onto the 2D frame and associated with 2D detections to determine object class assignment. All non assigned 3D objects are passed to the OctoMap module to update a static map, from which a 2D ground plane projection is used during dynamic path planning. Tracked 3D objects that have not yet been seen by the 2D camera but fit within a size threshold are included both as dynamic agents (shown in yellow in the bottom right) and static obstacles in the 2D occupancy map. }}
	\label{perception_pipeline}
\end{figure}

Multi-modal perception of static obstacles and dynamic agents in the robot's environment is achieved by combining 2D object detection in an RGB camera with 3D object segmentation and tracking from a \ac{LIDAR} point cloud. {Fig. \ref{perception_pipeline}} illustrates the steps involved in this process, resulting in classified and tracked 3D objects and a map of non-traversable static obstacles for use by the local dynamic planner.

\subsubsection{3D Segmentation and Tracking}

Input point cloud processing is performed to identify and track distinct objects in 3D. It also identifies traversable regions both for use by the \ac{FS} collision avoidance module and for generating a 2D occupancy grid, as shown in Fig. \ref{perception_pipeline}. This is performed in the following steps: (1) ground plane extraction; (2) segmentation of the point cloud into candidate `clusters’; (3) and tracking of the clusters in subsequent frames. 

Ground plane segmentation is achieved by initially thinning the input point cloud to 25\% before creating a 1 m grid voxel map. The lowest point in each voxel is determined using only the z-axis value, assuming the \ac{LIDAR} has been mounted upright and that any roll and pitch experienced by the robot reflects the local ground plane alignment. This assumption that sensor z-axis will be normal to the local ground plane is only valid for short distances when operating on non-planar ground. If the robot is on non-planar terrain, ground further away from the robot will be labelled as non-ground and only identified as being traversable as the robot approaches it and the robot’s z-axis approaches normal to the surface. This approach to ground segmentation has been chosen due to the restrictions of on-board computation, as opposed to more complex methods which allow non-planar ground, described in Section \ref{section:background_perception}.

The set of thinned points is iterated over, computing the height differences between each point and the lowest points in both its parent and the directly adjacent voxels. If this height difference is less than 0.2 m, the point is labelled as ground. The resulting non-ground point cloud set, shown in blue within the ground plane segmentation block of Fig. \ref{perception_pipeline}, is then passed both to the object segmentation and tracking block and the \ac{FS} collision avoidance module.
The non-ground points are then grouped into clusters using the clusters-all method described in \cite{Douillard_2011}. Points are partitioned only by local voxel adjacency, for which we use a local neighbourhood size of 0.3 m in x and y dimensions, and 0.5 m in z dimension, with a minimum cluster size of 20 points and minimum density of 2 points for each voxel. The segmented clusters are then tracked between frames based on centroid location. This is performed using Kuhn--Munkres association \cite{hungarian_method} and Kalman filtering \cite{Kalman1960} in 3D. Each track maintains a confidence score which is increased when the track is associated with a new detected cluster and decreased if not. If the confidence drops below a threshold, the track is deleted. This allows for the continuation of tracks when an object is briefly obstructed by other individuals or when a detection is otherwise missed. The output tracked-non-ground point set is then combined with 2D detections, as per Section \ref{section:fusion}.

\subsubsection{2D Object Detection}
Object detection in the camera frame is performed using the SSD: Single Shot MultiBox Detector \ac{CNN} \cite{Liu_2016}.
This network has been initialised using the the pretrained weightings from the VOC2012 dataset and then fine-tuned on a data set of labelled images relevant to agricultural applications. This dataset includes scenes from a number of Sydney University farms, as well as from the publicly available ImageNet dataset \cite{imagenet_cvpr09}, focusing on the classes of cows, sheep, horses and humans in order to improve detection performance in agricultural settings.
The output of the 2D detector---a series of classified 2D bounding boxes---is then combined with the output of the 3D segmentation and tracking module, described below.

\subsubsection{3D and 2D Fusion}
\label{section:fusion}
The output tracked-non-ground point cloud set is then associated with the output of the 2D object detector to determine class types by projecting the 3D point cloud onto the 2D camera frame. This step requires knowledge of the extrinsic transform between the two sensors and intrinsic parameters of the 2D sensor. The extrinsic transform is initially estimated, then refined using an unsupervised calibration between a camera and a \ac{LIDAR} as per \cite{Nebot2019}.  The 3D and 2D detections are associated by assuming that each detected 2D bounding box corresponds only to a single object, assigning the detected class label to the matched point cluster. This is determined by computing the intersection over union between the detected 2D bounding box, and a projection of the 3D detection onto the 2D plane. The Kuhn--Munkres algorithm \cite{hungarian_method} is again used to determine the best match for each detection.  Additionally, 3D bounding boxes are filtered using geometric thresholds based on the expected range of dimensions of the class to which the current 2D detection belongs.
Class confidence is updated for each assigned cluster based on output of the 2D \ac{CNN}.

The entire point cloud is then transformed into the world frame using the known robot transform, resulting in a geo-referenced point cloud segmented into ground, unknown-non-ground, and known-non-ground classes, with tracked centroid positions for each cluster within known-non-ground. The relative centroid positions of all tracked objects within the known-non-ground set, as well as all tracked objects within the unknown-non-ground set that pass the geometric threshold filter of the largest expected class, are then passed to the local dynamic planner. 

\subsubsection{Static Mapping}
\label{section:mapping}
The unknown-non-ground set is passed to the static mapping module in which a probabilistic OctoMap framework \cite{Hornung2013} is used to continuously update a map of static obstacles and traversable terrain. During each planning step, the output of this module is projected onto a 2D ground plane and resolved into a 60 $\times$ 60 grid of 0.5 m resolution centred on the robot's current position for use as an input to our local dynamic path planner, as shown in Fig. \ref{perception_pipeline}. 

\subsection{Long-term Planning}
\label{section:longplanner}
The long-term planner generates a strategic level mission plan that acts as the global reference path for the robot to track during operation. This mission plan is updated offline from sets of externally provided objective waypoints which the robot is required to visit in order to complete tasks such as weeding, soil sampling or---in the future---recharging. The \ac{PRM} algorithm is used to generate a traversability roadmap over free space $\mathcal{E}_{free}$, describing a set of kinematically feasible paths through the unstructured environment. This method proceeds in two phases: a learning phase, and a query phase. The roadmap generation occurs during the learning phase, capturing the traversability of the environment, as outlined in Alg. \ref{alg:prmgen}.

\subsubsection{Problem Definition}

The environment in which the robot will operate, $\mathcal{E} \subset \mathbb{R}^3$, is considered to be a 2D manifold embedded in 3D space, where $(x,y) \mapsto z$. Let the intraversible---or obstacle---regions of this environment be denoted $\mathcal{E}_{obs}$, such that $\mathcal{E} \setminus \mathcal{E}_{obs}$ is an open set. It is therefore implied that the freely traversible region of this environment is the closed set $\mathcal{E}_{free} = \text{cl}\left(\mathcal{E} \setminus \mathcal{E}_{obs}\right)$. Henceforth, let $\mathscr{I}^{n}_{m}$ denote the set of all integers from $m$ to $n$ inclusive, where $m \leq n : m, n \in \mathbb{Z}$.
\floatname{algorithm}{Algorithm}
\begin{algorithm}[!t]
	\caption{Probabilistic Roadmap Generation}
	\begin{algorithmic}[1]
		\Procedure {GenerateRoadmap}{$\mathcal{E}_{free}$, $\mathcal{U}$, $\rho_{PRM}$}
		\State $n_s, r_{conn}, v, \varepsilon_{min}, \varepsilon_{max} \leftarrow \rho_{PRM}$
		\State $\mathcal{V} \leftarrow \textsc{Sample}(\mathcal{E}_{free},n_s, r_{conn})$ \Comment{Sample $\mathcal{E}_{free}$ using chosen strategy}
		\State $\mathcal{V} \leftarrow \mathcal{V} \cup \mathcal{U}$ \Comment{Append goal nodes}
		\State $\mathcal{A} \leftarrow \textsc{GenEdges}(\mathcal{V},\ r_{conn})$
		\State $\mathcal{A}_{coll} \leftarrow \textsc{CollisionCheck}(\mathcal{A},\ \mathcal{E}_{free},\varepsilon_{min}, \varepsilon_{max})$
		\State $\mathcal{A} \leftarrow \mathcal{A} \setminus \mathcal{A}_{coll}$
		\State $\mathcal{M} \leftarrow \mathcal{V}, \mathcal{A}$
		\State $\mathcal{C} \leftarrow \textsc{CalcEnergyCost}(\mathcal{M})$ \Comment{Using ECM model}
		\State \textbf{return} $\mathcal{M}, \mathcal{C}$
		\EndProcedure
		\label{alg:roadmapgen}
	\end{algorithmic}
	\begin{algorithmic}[1]
		\Procedure {Sample}{$\mathcal{E}_{free}$, $n_s$, $r_{conn}$, $\mathscr{K}$}
		\State $\mathcal{V} \leftarrow \emptyset$
		\While {$\lvert \mathcal{V} \rvert < n_s$} \Comment{Iterative rejection sampling}
		\State $\chi \leftarrow \textsc{Rand}(1,\ \mathcal{E}_{free} \times \lbrack 0, 2\pi\rbrack)$
		\State $(\chi,z,\phi,\theta,\mathbf{C}) \leftarrow \textsc{RKP}(\chi,\ \mathscr{K})$
		\If {$\textsc{IsStable}(\chi,\mathbf{C})\ \land \neg \textsc{IsCollision}(\mathbf{C})$}
		\State $\mathcal{V} \leftarrow \mathcal{V} \cup \left( \chi \right)$
		\EndIf 
		\EndWhile
		\State \textbf{return} $\mathcal{V}$
		\EndProcedure
		\label{alg:kinesample}
	\end{algorithmic}
	\label{alg:prmgen}
\end{algorithm}
A set $\mathcal{V}$ of $n_s$ states $\chi_{i} \in \mathcal{V}: i \in \mathscr{I}^{n_s}_{1}$ are randomly sampled from $\mathcal{E}_{free}$---each state consisting of the 3DOF robot pose $(x,y,\psi)$---thereby discretising the continuous state space. Each sampling action involves solving an optimisation problem---namely, the relaxation of a 6DOF kinematic model of the robot onto $\mathcal{E}$---and the resultant pose is then checked both for collisions with $\mathcal{E}_{obs}$, and for static stability.

A path through the environment is defined by a continuous mapping $\zeta_{i,j} : \lbrack 0, 1 \rbrack \rightarrow \mathbb{R}^3$ such that $0 \mapsto \chi_{i}$ and $1 \mapsto \chi_{j}$. Each state $\chi_{i}$ is connected to its neighbours by paths in $\mathcal{E}_{free}$ to generate a roadmap $\mathcal{M} = \lbrace \mathcal{V}, \mathcal{A} \rbrace$, where $\mathcal{A} \subset \mathcal{V}\times\mathcal{V}$ is the set of arcs $a_{ij}\ \forall\ i,j \in \mathcal{V},\ i \neq j,\ \text{dist}(\chi_i, \chi_j) \leq r_{conn}$ connecting all vertices which are less than $r_{conn}$ metres away from each other, with $\text{dist}(\chi_i, \chi_j)$ representing here the 3D Euclidean distance between sample points $\chi_i, \chi_j$.

For each candidate connecting arc $a_{ij}$, the $\textsc{CollisionCheck}$ routine is invoked, where a minimum-curvature Clothoid curve connecting the two poses is generated, dilated by the maximum radial width of the robot, and subsampled along its length to check for both stability and collisions with $\mathcal{E}_{obs}$. If an unstable pose or collision is detected at any point along the path, or if the maximum curvature of the clothoid path exceeds a given threshold, the candidate arc is excluded from $\mathcal{A}$.

The long term energy efficient path planning problem can thus be denoted by the tuple $\left(\mathcal{M}^+, \chi, \chi_g \right)$, where $\chi_g = \mathscr{I}^{n_s+n_g}_{n_s+1}$ is the set of $n_g$ goal nodes, $\chi_{g,1}$ is the initial state, $\chi_{g,n_g}$ is the terminal or goal state, and $\mathcal{M}^+$ is the roadmap $\mathcal{M}$ augmented with the goal vertices and associated arcs connecting these to the roadmap.

By construction, a path $\pi$ through $\mathcal{M}^+$ will be feasible, and $\Sigma$ shall denote the set of all possible paths. The optimal path planning problem is therefore to find a path $\pi^*$, assuming $\left(\mathcal{M}^+, \chi, \chi_g \right)$ and an arc cost function $c : \Sigma \rightarrow \mathbb{R}_{\geq 0}$ such that $c(\pi^*) = \min \lbrace c(\pi) \rbrace$, or to report failure. The optimal path $\pi^*$ is considered to be $\delta$-robustly feasible if every point along the path trace is at least $\delta$ distance away from $\mathcal{E}_{obs}$.

As the learning process of the roadmap generation is an iterative one, it can be performed until an arbitrary number of samples are obtained of the environment. In more cluttered environments, for example, it may be desirable to sample densely to ensure feasible paths amongst the obstacles can be found reliably, whereas it may be desirable to sample more sparsely over large uncluttered environments to reduce the size of the roadmap, and thus the cost of querying it for paths.

\subsubsection{Solution Generation}

This roadmap $\mathcal{M}^+$ is subsequently queried using Dijkstra's algorithm, searching over the resultant graph to generate connecting routes between the provided objective waypoints. The minimum energy paths between all pairwise combinations of locations are determined using the energy cost of motion model developed in \cite{Wallace_WROCO} and the known topography of the environment.  

For extraction of the optimal motion plan from the roadmap, first let the goal connectivity graph $\mathcal{G} = \lbrace \mathcal{U}, \mathcal{L} \rbrace$ be defined as the graph encoding travel costs between goal nodes, where $\mathcal{U} = \mathcal{V}\setminus\mathscr{I}^{n_v}_{1}$ are the goal nodes, and the set of arcs $\mathcal{L} \subset \mathcal{U}\times\mathcal{U}$ are defined such that $l_{ij} \in \mathcal{P}_{ij,min} : i \neq j, i,j \in \mathcal{U}$, where $\mathcal{P}_{ij,min} \subset \mathcal{A}$ is the set of arcs describing the minimum cost path through $\mathcal{M}^+$ from $\chi_i$ to $\chi_j$.

$\mathcal{P}_{ij,min}$ is determined by querying the \ac{PRM}; performed by running Dijkstra's algorithm on $\mathcal{M}^+$ with start and goal points $\chi_i$ and $\chi_j$, respectively. All $\mathcal{P}_{ij,min}$ are stored along with their associated path cost $c_{ij} = \sum_{a_{ij} \in \mathcal{P}_{ij,min}} w_{ij}$ for later retrieval once the optimal tour $\mathcal{T}^*$ is found.

If the specified endpoint of the tour is not coincident with the start point, then the following arc weights are modified to enforce the precedence constraint: $c_{i1} = \infty : i \in \mathcal{U}\setminus\lbrace n_g \rbrace$, $c_{n_gi} = \infty : i \in \mathcal{U}\setminus\lbrace 1 \rbrace$, $c_{n_g1} = 0$. This assigns an infinite cost to all incoming edges to the start node, and all outgoing edges from the end node. Then, the edge connecting the end to the start node is given a weight of zero, and ignored in the final solution, to obtain a Hamiltonian path through the goals, where each vertex in the graph is visited exactly once. This procedure is outlined in Alg. \ref{alg:tourgen}.

An asymmetric traveling salesman problem is then solved over $\mathcal{G}$ to yield the optimal ordering of waypoint visits. The energy-optimal path $\mathcal{P}_{ij,min}$ is then extracted via reference to $\mathcal{M}^+$, thereby producing an energy-minimising plan suitable for use as the global reference path for the local planner.
\floatname{algorithm}{Algorithm}
\begin{algorithm}[!tb]
	\caption{Goal connection graph generation}
	\begin{algorithmic}[1]
		\Procedure {GenerateGoalConnGraph}{$\mathcal{M}$, $\mathcal{C}$, $\mathcal{U}$}
		\State $\mathcal{V}, \mathcal{A} \leftarrow \mathcal{M}$
		\State $\mathcal{P} \leftarrow \emptyset$
		\ForAll {$i,j \in \mathcal{U}$}
		\State $\mathcal{P}_{ij} \leftarrow \textsc{ShortestPath}(\mathcal{M},\chi_{g,i},\chi_{g,k})$
		\State $\mathcal{P} \leftarrow \mathcal{P} \cup \mathcal{P}_{ij}$
		\EndFor
		\State $\mathcal{L} \leftarrow \textsc{GenEdges}(\mathcal{U},\ \infty)$
		\State $\mathcal{G} \leftarrow \mathcal{U}, \mathcal{L}$
		\State \textbf{return} $\mathcal{G}, \mathcal{P}$
		\EndProcedure
	\end{algorithmic}
	\label{alg:tourgen}
\end{algorithm}
Through the use of Clothoid paths for connection of poses in $\mathcal{M}^+$, not only will the resulting motion plan by comprised of smooth, continuous motions, but by appropriate selection of the curvature rejection threshold parameter, it is possible to ensure that the path is feasibly trackable by a wide variety of vehicle classes. Combined with collision and stability checks along these paths, the feasibility of any resultant plans are guaranteed by construction. Further details of the above methods can be found in \cite{Wallace_Thesis2020}.

Replanning is also possible using this method, as the roadmap $\mathcal{M}^+$ is persistent, allowing alternate routes to be found by re-querying the roadmap in instances where the local planner has deviated significantly from the reference path. However, in the case where the environment state is inconsistent with the original map, or in cluttered environments where persistent dense crowds or similar phenomena inhibit progress along the nominal path, $\mathcal{E}_{obs}$ would need to be updated to reflect the intraversable area.
As recomputing the entire roadmap would be a very expensive operation to perform online, the existing roadmap can instead be pruned of any states and connections coincident with the new $\mathcal{E}_{obs}$, and then replanning can be conducted as normal on the modified roadmap. Replanning was not tested in the experiments conducted in this paper, but will be implemented as an item of future work.

\vspace{-0.1cm}


\subsection{Local Dynamic Planner}
\label{section:localplanner}
\begin{figure}[t]
    \centering
	\includegraphics[width=15.4cm,height=8.25cm]{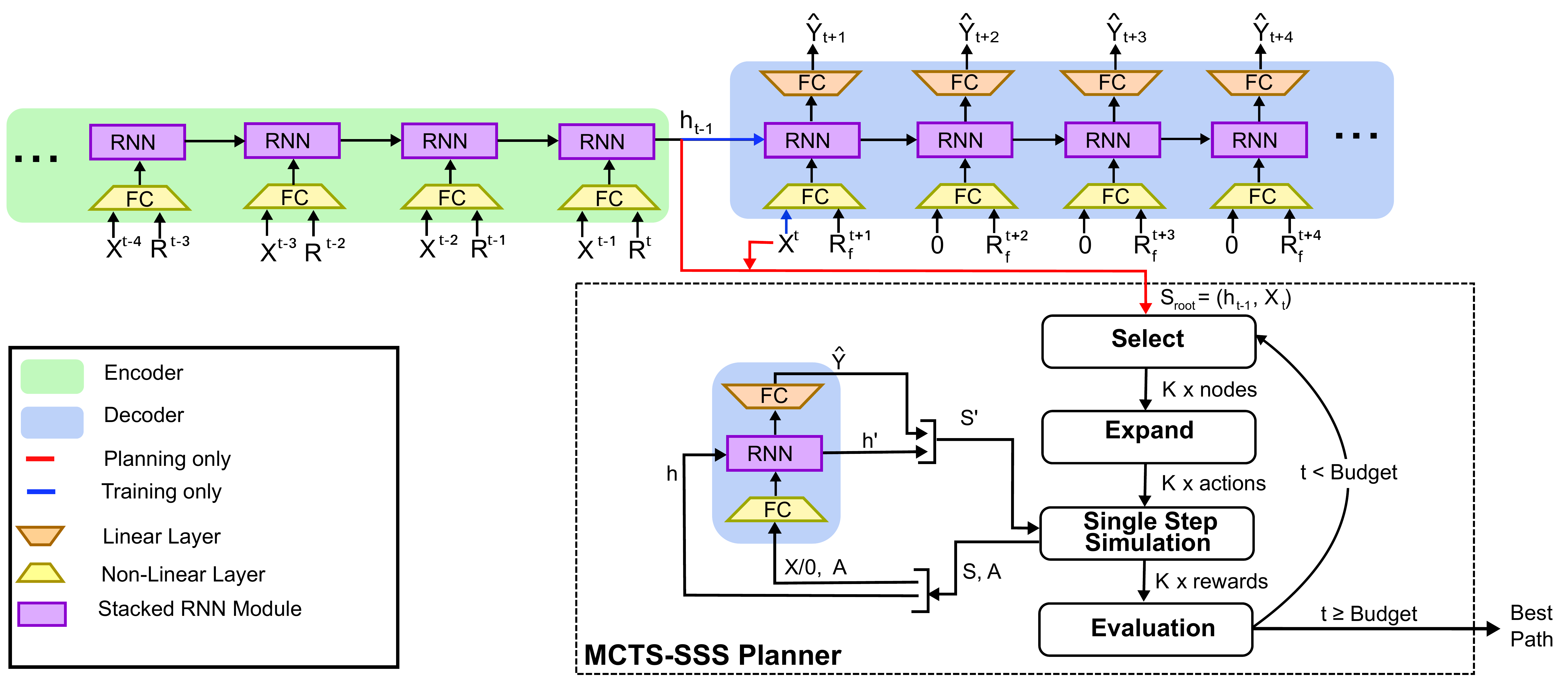}
	\vspace{0.5cm}
	\setlength{\belowcaptionskip}{5pt}
	\caption{\textit{\ac{MCTS}-\ac{GRNN} planner as per our prior work \cite{Eiffert2020a}, illustrating the integration of our learnt \ac{GRNN} predictive model of social response within the simulation stage of our adapted \ac{MCTS} Single Step Simulation planner.}}
	\label{mctsgrnn}
\end{figure}

The local planner is responsible for adapting the long-term plan based on the presence of both nearby moving individuals and unexpected obstacles. In order to do this effectively, an understanding of the relationship between the motion of nearby individuals and the robot motion is required, as discussed in Section \ref{section:background}.
As such, in this work our \ac{MCTS}-\ac{GRNN} version of the local dynamic planner is implemented, based on our prior work \cite{Eiffert2020a}. This planner uses a learnt \ac{GRNN} model of social response to predict the likely reactions to a robot's motion. This model is trained separately on observed robot-agent interactions and then used during a \ac{MCTS} across the robot's action space, as shown in Fig. \ref{mctsgrnn}. This allows simulation of the likely next state of a crowd for each sampled action. A state evaluation function, described in Section \ref{section:mcts}, is used to determine the value of each sampled action with respect to both reaching the robot's local goal and avoiding nearby agents. The sequence of actions with the highest cumulative value is returned by the planner each iteration, navigating the robot through a crowd towards the goal. 

The local planner module takes as input the tracked positions of all nearby agents, the current robot's state, a 2D occupancy map from the static mapping module, and the current local goal generated by the long-term planner, outputting a planned path to the hierarchical mode controller as illustrated in Fig. \ref{system_overview}. Whilst we only consider the use of a \ac{MCTS}-\ac{GRNN} local planner in the real-world trials carried out in this work, the local dynamic planner module within our proposed hierarchical framework is agnostic with regards to the used planner, as shown in our prior work \cite{Eiffert2020b} in which a \ac{PF}-based planner is implemented within the same framework for comparison in simulation. 

\subsubsection{Predictive Model}
\label{section:grnn}
The learnt model of social response used within our tested local dynamic Planner is based on a \ac{RED} baseline from \cite{Becker2018}.
The input to this model consists of a sequence of observations of an agent's position alongside the robot's known relative position in order to predict the agent's likely next position.
At each timestep, the robot's position at the subsequent timestep $R^{t+1}$ is used alongside the agents current position $X^t$, as shown in Fig. \ref{mctsgrnn}. This allows the model to learn the relationship between a robot's planned action and the likely response of an agent, thereby enabling its use as a state transition model within the local planner tree search.
At each planning timestep $t_{obs}$, the observed positions of all nearby agents from the prior 12 timesteps up to  $t_{obs}-1$ are used as input to the \ac{GRNN} encoder, shown in green. For any agents that have not been observed for the complete period, their history is extrapolated using a constant velocity model. We use a planning timestep of 200 ms in this work, based on the output rate of the perception module.


During training of the predictive model, the encoded output is passed to the decoder, where the known robot positions from the ground truth trajectory are used as input at each timestep $ t \geq t_{obs}$. The observed agent positions are also used as inputs at $ t = t_{obs}$, however, they are replaced by zeros in all subsequent decoding timesteps in a zero-feed approach to improve inference when the agent future positions are not available, as described by ~\cite{Zyner2018}.
When used for inference within the \ac{MCTS}-\ac{GRNN}, the output of the encoder is instead used to form the state of the root node within the tree search, and the decoder is used in a parallel single step manner within the \ac{MCTS} simulation step, as discussed in Section \ref{section:mcts} below.

We make use of the model trained on the Vehicle-Crowd Interaction (VCI) DUT dataset ~\cite{Yang2019} of human vehicle interactions in our prior work \cite{Eiffert2020a}, as this model best reflects the use case of the experiments carried out in this work, in which a wheeled mobile robot navigates around pedestrians.
In \cite{Eiffert2020a}, we have also shown how our \ac{MCTS}-\ac{GRNN} planner can be used when a predictive response model is not available, by substituting a simple constant velocity model in place of the \ac{GRNN}. Whilst this approach is unable to model the relationship between a robot's action and the likely response of a crowd, it is still able to achieve results that outperform the compared \ac{PF} method.

\subsubsection{Sampling-based Planner}
\label{section:mcts}
Alg. \ref{alg:mcts} details the steps involved in the \ac{MCTS}-\ac{GRNN} response-aware planning module used within this work. This anytime planner searches across a discretised action space of the robot, taking as input a known state evaluation function. Each iteration, the latest observation of nearby agents is used alongside the robot's current state and a 2D occupancy map as per line 26 in Alg. \ref{alg:mcts}. The tree search concludes after a given time budget has been reached, returning the current best plan.

\begin{algorithm}[t]
	\setlength{\belowcaptionskip}{-5pt}
	\caption{\strut \ac{MCTS}-\ac{GRNN} Planner}\label{alg:mcts}
	\begin{algorithmic}[1]
		\State A $\gets$ Actions \Comment{\textit{discretised action space}}
		\State B $\gets$ Budget \Comment{\textit{time in nsecs}}
		\State C $\gets$ CostFunction() \Comment{\textit{state evaluation function}}
		\State M $\gets$ Map \Comment{\textit{2D Occupancy Map}}
		\Function{MCTS-SSS(root, A, B, C, M)}{}
			\State Tree = createTree(root) \Comment{\textit{Create Tree with root state as first node}}
			\While{\texttt{time < B}} \Comment{\textit{planning budget}}
				\State K = Tree.select(root) \Comment{\textit{select K best nodes}}
				\State a = Tree.expand(K, A, M) \Comment{\textit{choose valid actions}}
				\LineComment{\textit{ parallel single step simulation}}
				\If {first iteration}
					\State $h',\hat{Y'}$ = RNNDecoder($X^t$, a, h)
				\Else
					\State $h',\hat{Y'}$ = RNNDecoder(0, a, h)
				\EndIf
				\State U = $\sqrt{det(cov(\hat{Y'}))}$ \Comment{\textit{uncertainty}}
				\State r = C($\hat{Y'}$,U ) \Comment{\textit{reward dependent on U}}
				\State Tree.backup(K, r) \Comment{\textit{update node values}}
			\EndWhile\label{euclidendwhile}
			
			\Return{\texttt{Tree}}
		\EndFunction
		\While{\texttt{not at destination}}
			\State $X^{0:t}, R^{0:t-1}$ $\gets$ observe() \Comment{\textit{Positions of nearby agents X and robot R for past t timesteps}}
			\LineComment{ \textit{ encode observed tracks}}
			\State $h^t,\hat{Y}^t$ = RNNEncoder($X^{0:t-1}, R^{0:t-1},h^0=0$)
			\State $S_{root}^t = (h^t,X^t)$ \Comment{\textit{create root node}}
			\LineComment{ \textit{ perform MCTS with SSS}}
			\State Tree = MCTS-SSS($S_{root}^t,A,B,C,M$)
			\State $R_p$ = Tree.bestPlan \Comment{\textit{yield best current path}}
		\EndWhile\label{euclidendwhile}
	\end{algorithmic}
\end{algorithm}

The used \ac{MCTS} has been adapted for single step simulation. This involves running the simulation stage for only a single iteration rather than continuing until a terminal state has been reached, as shown at line 8 of Alg. \ref{alg:mcts}. As described in \cite{Eiffert2020a}, this allows parallelisation of the simulation step by ensuring all rollouts are of the same length. Similar truncation of the simulation stage has been shown to improve performance in game-theoretic applications such as GO \cite{Silver2016} and Amazon ~\cite{Lorentz2016}, when the value of a state can be directly evaluated.

The state evaluation function used during the tree search is shown in Eq. \ref{eqn:costfn}. This cost is dependent on the distance of the robot to the local goal $G$, and the scaled distance $\alpha$ between the robot and each agent $X_i^t$ for all observed agents $i\in N$ at the current timestep $t$. The uncertainty $U_i^t$ of the prediction for agent $i$ is used to scale $\alpha$, and is set to zero when the distance between robot and agent exceeds a value of $d$, which is set to 2 m for all trials. $U_i^t$ reflects the area of the ellipse formed by 1 standard deviation from the mean prediction. The major radius of this ellipse can in theory exceed the value of d, meaning that an agent may be predicted to come within the distance threshold without any cost being considered. However, as the uncertainty tends to grow with prediction horizon length, this would likely only occur at longer prediction timesteps and has not been observed to occur at timesteps less than 3 s, and so not significantly influence the MCTS planner.

\begin{flalign}
Cost &= (R^t - G)^2 + \sum_{i}^{N}  {U_i^t} \alpha\label{eqn:costfn}\\
\alpha&= 
\begin{cases}
    \frac{1}{X_i^t - R^t} ,& \text{if } X_i^t - R^t\leq d\\
    0,              & \text{otherwise}
\end{cases}\label{eqn:alpha} 
\end{flalign}


A 2D occupancy map centred on the robot, as shown in Fig. \ref{perception_pipeline}, is used to constrain the action space. During \ac{MCTS} expansion only actions which do not result in a collision between the robot and the occupancy map are allowed. The occupancy map is also dilated to the maximum radius of the robot to prevent the problem of static obstacle `corner cutting' in sampling-based planners, as described in \cite{Solovey2020}, in which a collision may occur between two discrete timesteps.
\vspace{-0.1cm}


\newpage
\subsection{Hierarchical Mode Switcher}
\label{section:highlevelcontrol}

\floatname{algorithm}{Algorithm}
\begin{algorithm}[t]
	\setlength{\belowcaptionskip}{-5pt}
	\caption{\strut Hierarchical Mode Switcher }\label{alg:modeswitch}
	\begin{algorithmic}[1]
	    \State waypoints $\gets$ External Mission Objectives
	    \State LT $\gets$ LongTermPlanner() 
	    \State LT.computeOptimalPath(waypoints)
	    \State LDP $\gets$ LocalDynamicPlanner() 
	    \State PathTracker $\gets$ RecedingHorizonEstimatorController() 

		\While{\texttt{not at LT.terminalWaypoint}}
	    	\While{\texttt{not at LT.currentWaypoint}}		
				\If {FailSafe.active}
					PathTracker.stop()
				\Else
    				\If {LDP.required} \Comment{\textit{Check dynamic planning area}}
    					\LineComment{\textit{If we expect a planning delay, immediately slow and get expected robot state after delay}}
    				    \State expectedState = PathTracker.slow(LDP.planningDelay) 
			            \State latestDynamicPlan = LDP.getPlan(expectedState) \Comment{\textit{Plan from expected end state}}
    				    \State dynamicLatch.resetTimer() \Comment{\textit{Reset latching timer}}
                    \EndIf
    				\If {dynamicLatch.timer $<$ 2 s} \Comment{\textit{Avoid oscillating between modes}}
    					\State path = latestDynamicPlan \Comment{\textit{Use latest dynamic path}}
    				\Else
    					\State path = LT.localGoal \Comment{\textit{Use latest long term plan local goal update directly}}
				    \EndIf
    			\State PathTracker(path) \Comment{\textit{Send to Path Tracking module}}
				\EndIf
			\EndWhile\label{euclidendwhile}
    		\State  PathTracker.stop()
    		\State 	Weeder.actuate() \Comment{\textit{Stop and weed}}
		\EndWhile\label{euclidendwhile}

	\end{algorithmic}
\end{algorithm}

Based on the robot's proximity to detected dynamic agents and static obstacles, a hierarchical mode switching module---detailed in Alg. \ref{alg:modeswitch}---determines whether to source the local reference trajectory from the dynamic planner, or to follow the online update of the global path provided by the long-term planner. This represents a crucial element in the integration of the global optimal planner and the local dynamic planner within this framework.

In the absence of obstacles, the default path tracking behaviour will compute a reference based on the robot's progress along the global path and the specified nominal speed. If a dynamic agent or static obstacle is detected within the dynamic planning area, the local dynamic planner will then be engaged. Depending on the type of local dynamic planner being used, there will be up to a 200 ms planning delay between detection of the need to dynamically plan and the plan being finalised. In this time, the hierarchical switcher will send an update to the RHEC tracker to decrease velocity over the next 200 ms (or known planning delay) along the current path. The expected new velocity and future position are passed to the dynamic planner to use as the state from which to begin planning. In the case of the MCTS-GRNN planner described in Section \ref{section:localplanner}, this requires predicting the response of nearby agents to the robot's expected state for use in the \ac{MCTS} root node. In the case of \ac{SARL}, which we compare as an alternative planner in Section \ref{section:experiments}, we simply propagate agent positions forward using the same constant velocity model used to propagate future states passed to the value network \cite{chen2019crowd}.  

The dynamic planning mode is latching for 2 s, ensuring that the hierarchical switcher does not rapidly oscillate between dynamic and long term modes when an obstacle is on the edge of the dynamic planning area resulting in unstable behaviour. This area is defined as the union of a forward facing semicircle 4 m in radius centred on the robot's centre, and a similarly forward facing circular sector with subtended angle of 135$^\circ$ and radius of $ (4+2v)$ m, where $v$ is the robot's current speed.



Due to possible planning delays of the local planner, we also make use of a \ac{FS} collision avoidance module, ensuring that the robot is able to reflexively react to rapid changes in its environment without having to wait for the local planner to complete planning. This module simply stops the robot in case of a potential collision and waits for the path to clear, rather than planning a path around obstacles. It makes use of the \ac{LIDAR} input directly after ground plane segmentation and removal, as shown in Fig. \ref{perception_pipeline} for faster reaction time. To determine if a collision is imminent, the \ac{FS} module checks a safety area in front of the robot for obstacles in a similar manner to the dynamic planning area, however with a semicircle radius of 2 m and circular sector defined by an angle of 90$^\circ$ and radius of $ (2+v)$ m.


\section{Experiments}
\label{section:experiments}

The proposed planning framework has been evaluated in both real-world and simulated experiments. Simulated trials, building upon the analysis presented in \cite{Eiffert2020b}, have been used to allow comparison of varied local dynamic planners within the hierarchical framework during extended navigation. Subsequent real world trials, building upon our prior trial in \cite{Eiffert2020c}, have been used to validate performance on physical hardware deployed in a real operating environment.



\begin{figure}[t]
    \centering
	\includegraphics[width=12.0cm,height=9.0cm]{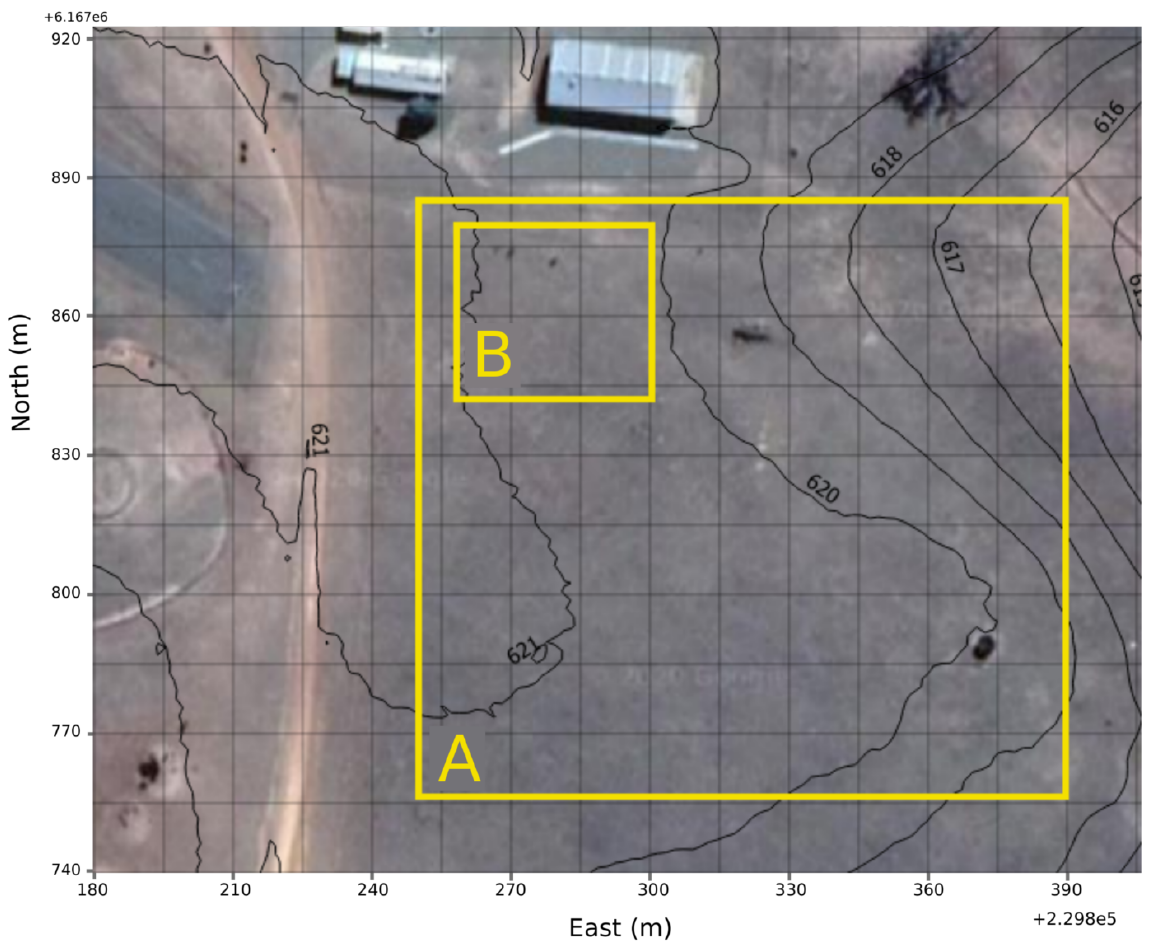}
	\setlength{\belowcaptionskip}{0pt}
	\caption{\textit{Aerial map of the University of Sydney's Arthursleigh Farm used for all real world trials, illustrating the location and topography of the  extended navigation trial  (a) and the crowd interaction trial  (b).}}
	\label{overview_map}
\end{figure}

During simulated trials, the framework has been compared in varying densities of agent crowds, and when using different local dynamic planner modules. Trials in \cite{Eiffert2020b} compared the use of a \ac{MCTS}-\ac{GRNN} approach, a \ac{PF}-based approach, and when relying only on the \ac{FS} collision avoidance module. This work has expanded the trials to include comparison when using a state of the art reinforcement learning planner as the local dynamic planner, \ac{SARL} \cite{chen2019crowd}.
The real-world testing of our approach was conducted on the University of Sydney's Swagbot agricultural robot platform, shown in Fig. \ref{robotic_platform}. Testing was performed at the University's Arthursleigh Farm and involved two separate trials. 
The first trial, detailed in \cite{Eiffert2020c}, involved extended navigation between mission waypoints across an unstructured field 2 ha in size for the purposes of weeding, whilst in the presence of dynamic agents and unknown obstacles. The second trial focused exclusively on the robot's interaction with dynamic agents and was conducted on a smaller scale with denser crowds in order to comprehensively evaluate the behaviour of our proposed framework during crowd and herd interactions. An overview of the real world trials is shown in Fig. \ref{overview_map}, detailing the extent of the extended navigation trial (a) and crowd interaction trial (b). Additionally, an evaluation of the perception pipeline's ability to accurately detect the location of nearby agents was also carried out. This was performed using a comparison between the output of the object detection and tracking module and footage from an overhead drone, described in Section \ref{section:perception_analysis}.


\subsection{Experimental Platform}
\label{hardware}
\begin{figure}[H]
    \centering
	\includegraphics[width=15.5cm,height=7.0cm]{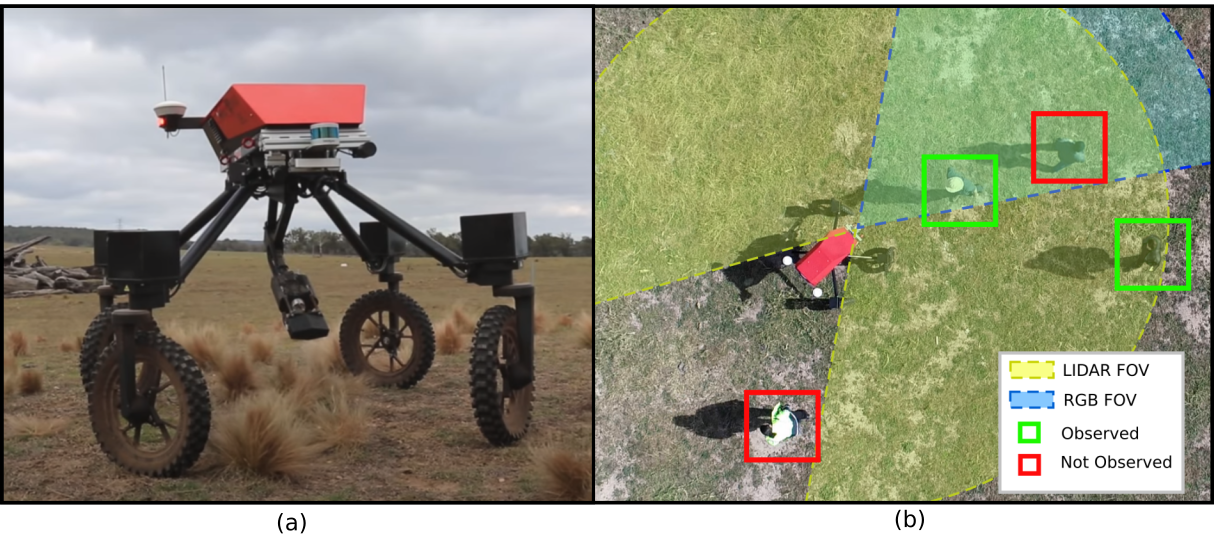}\vspace{-0.2cm}
	\setlength{\belowcaptionskip}{-10pt}
	\caption{\textit{University of Sydney's Swagbot Agricultural robotic platform used in all real-world testing. (a) Photo of Swagbot with actuated weeder extended at the extended navigation trial location. (b) Top-down illustration of sensor \ac{FOV} with obstruction of the \ac{LIDAR} by the robot's chassis, demonstrating missed detections resulting from both sensor blindspot and crowd occlusions.}}
	\label{robotic_platform}
\end{figure}

The Swagbot robotic platform was used in all trials conducted during this work. This platform is a wheeled omnidirectional electric ground vehicle designed for use in uneven terrain such as grazing livestock farms. The platform has a limited battery capacity of 1.97 kWh with expected drive time of approximately 3 hours before requiring recharging.
As shown in Fig. \ref{robotic_platform} (a), the robot includes an actuated arm attached to the underside of the chassis, intended for use in tasks such as weed spraying and soil sampling. 
The localisation system includes a Trimble BD982 \ac{GNSS} Receiver and Orientus V3 \ac{IMU}, which provide estimates of the position and orientation of the robot. A forward-facing Point Grey Grasshopper GS3-U3-51S5C-C and Velodyne VLP-16 \ac{LIDAR} are mounted on the front of the robot for use in obstacle detection and mapping. The configuration of these sensors results in a limited \ac{FOV} and rear blind spot, as shown in Fig. \ref{robotic_platform} (b). Additionally, a downwards facing RealSense D435 camera is mounted below the \ac{LIDAR} for use in active perception during weeding. As the accuracy of the active perception system was not evaluated in this work, this camera was not utilised during the trials.
\begin{figure}
    \centering
	\includegraphics[width=15.0cm,height=6.8cm]{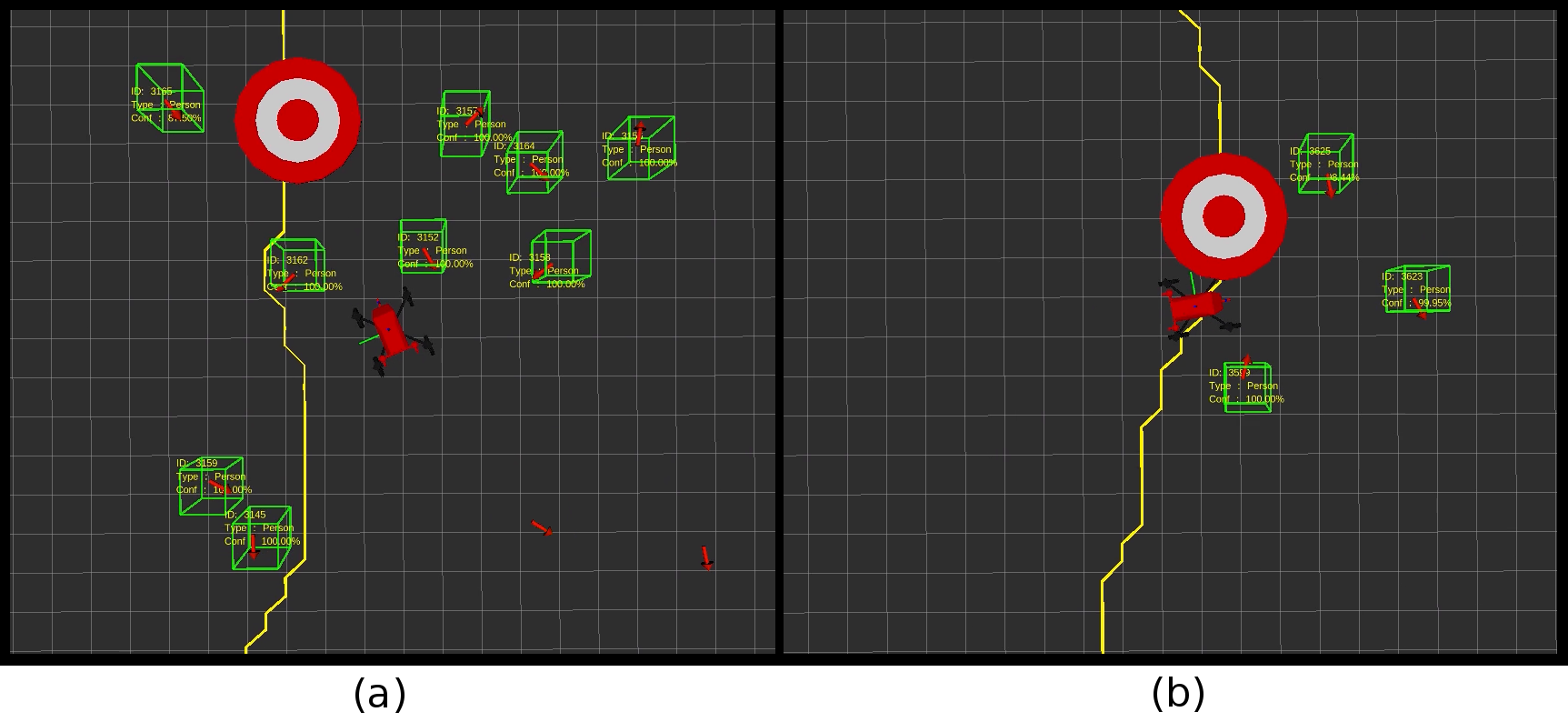}
	\setlength{\belowcaptionskip}{-5pt}
	\caption{\textit{Simulated environment used for testing, illustrating crowd densities of 10 $m^2$ (a) and 25 $m^2$ (b) per agent on a $1\times1$ m grid. The long-term planner's reference path is shown in yellow with the current waypoint as a target. Simulated \ac{ORCA} agents are shown as red arrows, with green bounding boxes highlighting agents that the robot is currently tracking.}}
	\label{sim_env}
\end{figure}

\subsection{Methodology}
\subsubsection{Simulated Trials}
\label{simtrials}
Our prior work \cite{Eiffert2020b} describes the simulated testing, in which the robot was required to navigate between a set of mission waypoints through an unstructured environment, taking into consideration the presence of nearby individuals and a limited energy budget. Real-world aerial terrain data from the University of Sydney's farm at Bringelly was supplied alongside mission waypoints to the robot in order to compute an offline reference path as per Section \ref{section:longplanner}. Between 5-12 waypoints were supplied to the robot each iteration, with average spacing of 25 m between each. The robot was required to visit each waypoint within the resource budget. Perception of dynamic agents and localisation was simulated to reflect real-world sensors available on the Swagbot platform, including sensor noise and \ac{FOV} limitations. The simulated environment is shown in Fig. \ref{sim_env}, illustrating simulated agents as red arrows, with detected agents shown in green boxes.

This simulated trial was repeated using four different framework implementations, in which the local dynamic planning module was using either: (1) \ac{MCTS}-\ac{GRNN}-based planner described in Section \ref{section:localplanner}; (2) \ac{SARL} \cite{chen2019crowd}, a state of the art reinforcement learning based dynamic path planner which uses attention based mechanisms to capture both robot-agent and agent-agent based interactions; (3) a \ac{PF}-based approach as per our prior work \cite{Eiffert2020a}; or (4) \ac{FS} collision avoidance only, with no dynamic planner, as shown in Fig. \ref{system_overview}.

This was repeated 3 times per planner version with varying required positional accuracy at the mission waypoints, of 5 m, 2 m and 1 m.
Additionally, this testing was conducted in two different agent crowd densities, of 10 m$^2$ (dense) and 25 m$^2$ (sparse) per agent, illustrated in Fig. \ref{sim_env}. Each test was undertaken in real-time, taking approximately 2.5 hours to reach all supplied waypoints, depending on the type of dynamic planner and agent density used. Agents were simulated in these trials using the \ac{ORCA} \cite{VanDenBerg2011} pedestrian motion model with speeds between 0.1--1.5 m/s and maximum neighbour distance of  1.5 m. This term has been edited from the original implementation  \cite{VanDenBerg2011} to refer to the distance between agents excluding radii, rather than centre to centre distance, to accommodate heterogeneous agent sizes. All simulated trials used an agent radius of 0.5 m, and a robot radius of 1.5 m.

 \begin{figure}[H]
    \centering
	\includegraphics[width=16.0cm,height=8.0cm]{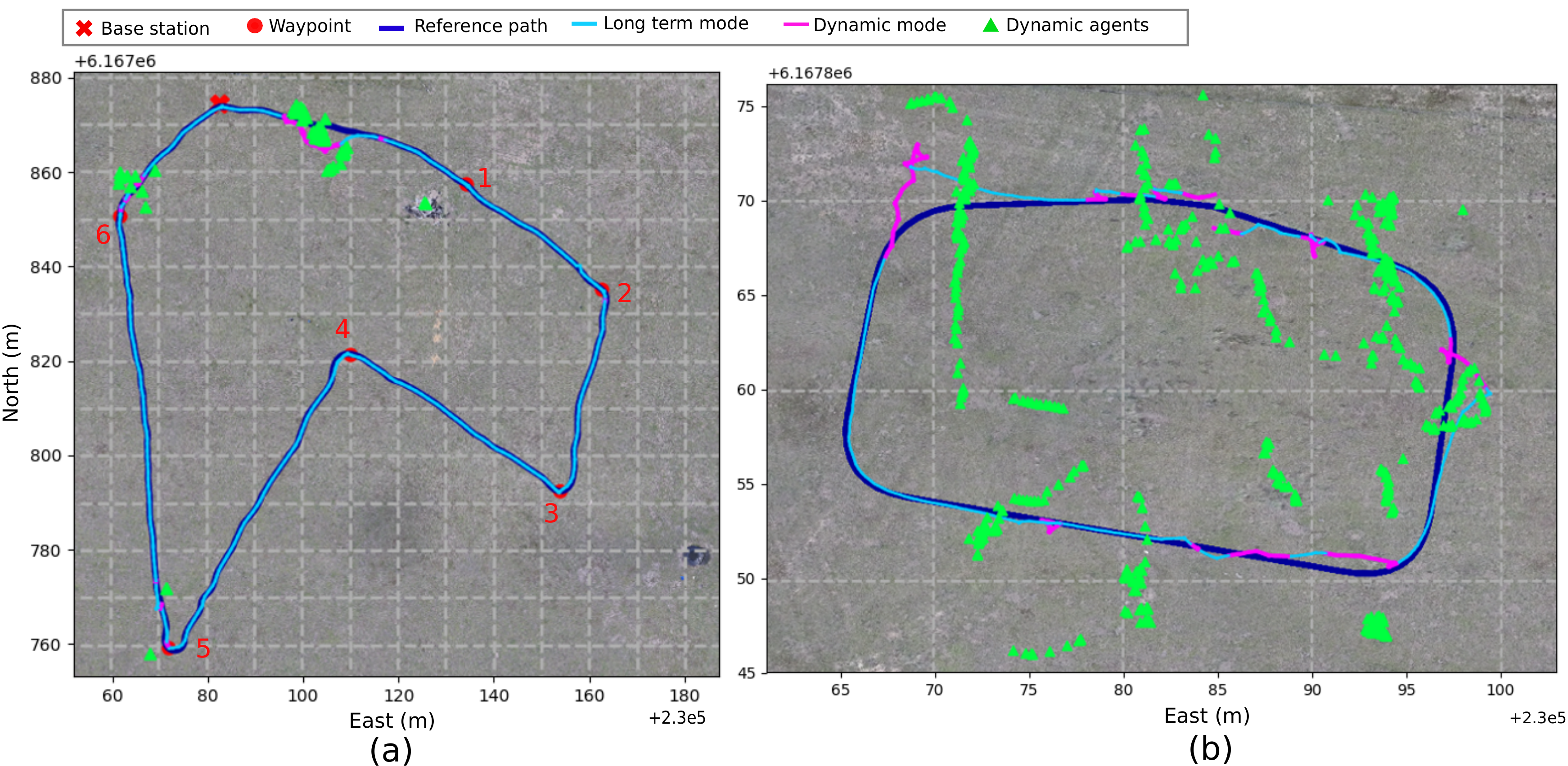}
	\setlength{\belowcaptionskip}{-0pt}
	\caption{\textit{Example iterations of the extended navigation trial (a) and the crowd interaction trial (b) showing differences in scale, agent density, and relative time spent in each planning mode. Note that the right map is shown at 1:4 scale of the left. The reference path (dark blue) is shown alongside the actual taken path, differentiating between when the long-term mode (light blue) and dynamic mode (purple) were each being used. The location of all detected agents throughout each trial lap are shown as green triangles.}}
	\label{trial_map2}
\end{figure}
\subsubsection{Extended Navigation Trial}
The real-world extended navigation trial aimed to replicate the methodology used in the simulated testing, involving continuous navigation between updated sets of externally provided mission waypoints across an unstructured pastoral field. At each waypoint the robot was required to reach a positional accuracy of 1 m in order to spray a pre-located weed, operating in the presence of both moving individuals and unknown obstacles. Each iteration of the trial began at a base waypoint, where a set of 5-8 objective waypoints were supplied to the robot.

An offline resource-efficient path was again determined based on a prior terrain map generated from aerial \ac{LIDAR} survey, and used as the reference path for online local planning during navigation to each objective. 
Upon returning to base, a new set of waypoints were supplied and the trial repeated. A total of 3 sets of waypoints were reused, with testing continuing until the robot exhausted its energy resources. An overview map of the testing area, showing a single example iteration of the extended trial, is illustrated in Fig. \ref{trial_map2} (a). The hierarchical planning framework was implemented using the same \ac{MCTS}-based local dynamic planner used in simulated trials and the perception pipeline as described in Section \ref{section:perception}. 
Total time of the trial was 2 hr 44 mins, covering a distance of 5.49 km, including 37 separate interactions with groups of moving agents.

\subsubsection{Crowd Interaction Trial}
The second real-world trial involved the continuous tracking of a reference path whilst navigating through a sparse crowd of pedestrians. The reference path was a circuit of approximately 85 m in length as shown in Fig. \ref{trial_map2} (b), computed using the long-term planner but not requiring the robot to stop at any waypoints. This trial used the same \ac{MCTS}-based local dynamic planner as the extended navigation trial. Eight pedestrians were involved, who were instructed to begin outside the perimeter of the robot's reference path and to choose a goal point on the other side of the circuit which they aimed to walk towards. Upon reaching their goal, pedestrians would again choose another point, continuing back and forth for the duration of the trial. Pedestrians were not instructed to give way to the robot, instead being told only to treat the robot as if it were being driven by a human operator. This trial lasted a total of 21 minutes, in which time the robot completed six laps of the reference circuit.  
\subsubsection{Perception Pipeline Evaluation}
\label{section:perception_analysis}
An evaluation of the accuracy of the object detection and tracking module was carried out during the crowd interaction trial. This involved a comparison of the classified and tracked 3D objects output by the perception pipeline to the ground truth positions of all nearby agents, obtained from an aerial drone video.
Aerial images were labelled at 24 fps using initial manual detection of both the robot's location and each pedestrian in the 2D aerial image. Kernelized correlation filter \cite{Henriques2015} based visual tracking was used to automatically label subsequent frames, with manual re-initialisation of tracks as required. 2D pixel position of each pedestrian in polar coordinates, relative to the robot's location, was saved each frame.

\begin{figure}[t]
    \centering
	\includegraphics[width=9.0cm,height=8.0cm]{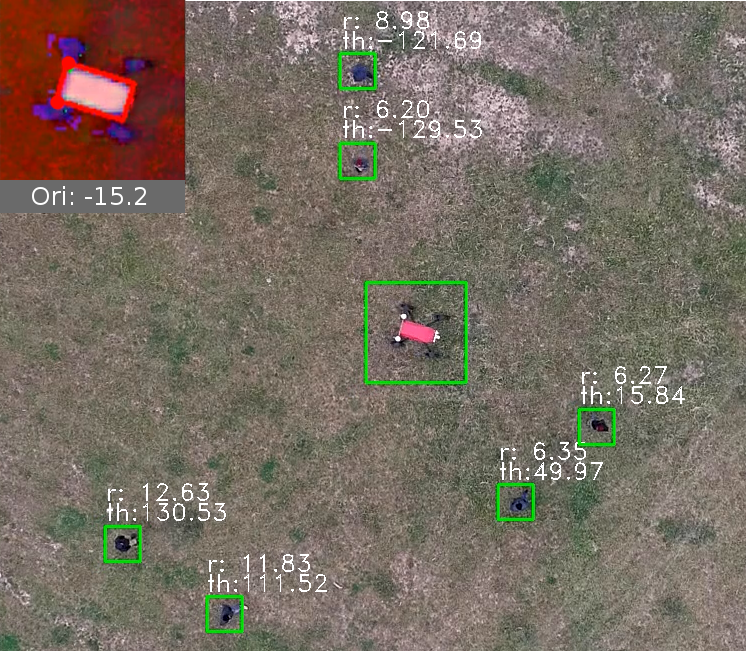}
	\setlength{\belowcaptionskip}{-0pt}
	\caption{\textit{Labelled overhead image used as ground truth for perception pipeline evaluation. Robot and agent locations are tracked in green, with scale and robot orientation being determined by location and size of red robot chassis and white antenna of known geometry in each frame, as identified in the inset. }}
	\label{labelled_drone}
\end{figure}

Tracked positions were then transformed into global scale relative to the robot's orientation. Global scale was estimated each frame using the known geometry of the robot and visible features on the robot as fiducial markers. This included identifying the outline of the red chassis and the location of the two white aerial enclosures each frame through the use of colour based thresholding and contour detection within the tracked 2D bounding box of the robot. These features are visible in the top-down image Fig. \ref{robotic_platform} (b). These centres of each marker were tracked between frames using Kalman filtering \cite{Kalman1960}, and used to determine both the orientation of the robot, and the pixel-to-metre scaling for each frame. An example labelled frame is shown in Fig. \ref{labelled_drone}, illustrating both the tracked 2D boxes and scaling markers.

Comparison of the ground truth labels to detection output was done at 2 Hz using a total of 2420 frames. Each ground truth labelled position was marked as `detected' if a detection was made within a 1 m radius of the label with up to 0.5 s difference in timestamp. Section \ref{perceptionresults} summarises the results of this trial, describing the recall and precision of the perception pipeline across the robot's sensing space.

\subsection{Metrics}

Performance in all trials, both real-world and simulated, has been compared based on metrics of: (1) distance to closest agent; (2) energy usage per metre gained towards the goal; (3) velocity towards the goal; and (4) deviation from the reference path. Metrics 2, 3 and 4 have all been calculated in varying crowd densities to determine the impact of more complex environments on resource efficiency.

Metric 1 represents the ability of each planner to effectively navigate through crowds, providing a measure of the safety of the system around moving individuals and the number of potential near-collisions.
Metric 2 represents the energy expenditure of each trial, and has been normalised for comparison between trials as J/m gained towards goal. This was chosen over the average energy usage per trial as it allows direct comparison of energy usage in varying crowd densities. Power usage is calculated using the learnt energy cost of motion model outlined in Eq. \ref{eqn:ptmass2}, derived below:

\begin{align}
P &= F \cdot v + \sum\limits_{s \in \mathcal{S}}\varsigma_{s}(t) \label{eqn:pmmgencont}\\
&= F \cdot v + \sum\limits_{s \in \mathcal{S}}q_{s}\varsigma_{s} \label{eqn:pmmgen}\\
&=  \left(m_ig \sin\left(\varepsilon\right) + \mu_i N \right) \cdot v + \sum\limits_{s \in \mathcal{S}}q_{s}\varsigma_{s} \label{eqn:ptmass}\\
&= \left(\sin\left(\varepsilon\right) + \mu_i\cos\left(\varepsilon\right) \right) \cdot m_igv + \sum\limits_{s \in \mathcal{S}}q_{s}\varsigma_{s} \label{eqn:ptmass2}
\end{align}

where $P$ is the instantaneous power draw of the robot, $F$ is the force acting in the direction of motion, $v$ is the speed of the robot in meters per second, $\varepsilon$ is the slope of the terrain along the robot's path of motion in radians, $m_i$ is the mass of the platform at the time of run $i$, $g$ is the acceleration due to gravity, $N$ is the normal force acting perpendicular to the ground, $\mu$ is the coefficient of rolling friction at the site of the run, $\varsigma_{s}(t)$ and $\varsigma_{s}$ are the dynamic and static power draw of subsystem $s\in \mathcal{S}$, respectively---where $\mathcal{S}$ is the set of computers, sensors and actuators---and $q_{s} \in \lbrace 0,1\rbrace$ is a binary variable with a value of 1 if a given subsystem is active, or 0 if it is not. For the experiments conducted in this work, $m_i = 220.6$ kg, $\mu_i = 0.0767$, and the static power draw of the system is $\sum\limits_{s \in \mathcal{S}}q_{s}\varsigma_{s} = 203$ W. Further details can be found in \cite{Wallace_WROCO}. 

Metric 3 represents the navigation time efficiency of each trial. Similarly to metric 2, velocity towards goal was chosen over total experiment time ---  a more intuitive metric for time efficiency --- as it can be determined at every timestep to allow comparison in varying agent densities. For the analysis done in Section \ref{resourcenavigationresults}, the average velocity across every 1 second period during each trial was used against the maximum number of agents detected within the same period.

Metric 4 represents the ability of the planner to accurately follow the energy efficient path, a crucial consideration in environments with large elevation changes that may lead to significant energy usage to return to the reference path after a minor deviation.

\subsection{Implementation}

During the real-world trials carried out in this work, the output of the static mapping module was not utilised by the local dynamic planner. This was done for clarity, allowing testing to focus on the performance of the planning framework around moving individuals only, and relying on the \ac{FS} collision avoidance module to ensure safety around static obstacles.

Due to the limited sensor \ac{FOV}, as indicated in Fig. \ref{robotic_platform} (b), the robot was constrained to operate in forward-only Ackermann configuration during the simulated tests. This was done to restrict motion in directions without adequate perception coverage for safety precautions. In the real-world trials, an alternate approach was used, where this \ac{FOV} safety constraint was instead enforced in the local path planning step, which would only generate paths in the region covered by the robot's sensors. This change was taken after observing that operation in forward-only Ackermann configuration often made direct and precise visitation of goal locations in the presence of numerous moving agents difficult, due to the mobility constraints it imposes.

The nominal path tracking module used in the simulated tests utilised a slip-compensating \ac{RHEC} framework---which uses model-based localisation and predictive control to minimise cross-track and speed tracking error \cite{wallace2019b,wallace2019a}. This strategy utilised a forward-only Ackermann motion model. In the real-world tests, however, a pure pursuit path tracker in conjunction with a holonomic motion model was used instead; employing a \ac{PID} controller to drive the robot's positional error relative to a moving target point to zero via linear and angular velocity control of the motion base. The change in path tracking strategy for the real-world runs was chosen as it did not artificially restrict the holonomic motion capabilities of the platform---as the Ackermann-based \ac{RHEC} strategy would have---thereby allowing more direct motion towards the goal locations to be realised. Both trackers otherwise worked to maintain constant speed, and control orientation to face the direction of travel along the reference path.

During operation, the long-term planner outputs an online local goal which  moves along the computed global reference path 10 m in front of the robot's position towards the next waypoint. This local goal is tracked by either the local dynamic planner, described in Section \ref{section:localplanner}, or directly by the nominal path tracking module described above, depending on the mode dictated by the hierarchical mode controller module. The local planner additionally generated plans which forbid reversing, in order to restrict the robot from moving into areas in its blindspot.
For generation of the long-term energy-efficient path plans, the \ac{PRM} strategy introduced in Section \ref{section:longplanner} was used for roadmap generation, using a connection radius $r_{conn} = 4$ m and a maximum curvature constraint value corresponding to a max Ackermann steering angle of $35^\circ$.

For the purposes of all trials in this work, pedestrians---rather than livestock---were used as dynamic agents. This was decided for a number of reasons, including compatibility with simulated trials, available response prediction models, and safety of agents. As discussed in \cite{Eiffert2019}, the use of  mobile robots around livestock is not as widespread as for pedestrians or traffic, with little work done to explore how best to model livestock motion, especially in response to a robot. Whilst \cite{Eiffert2019} was able to show improved prediction accuracy on the livestock ARATH dataset \cite{JUnderwood2013} using spatio temporal graph RNNs (STGRNN) compared to constant velocity models, later work \cite{Eiffert2020a} showed that when using the simpler GRNN model used in this work --- required for use in the MCTS-GRNN due to faster inference speed than STGRNN --- the motion prediction accuracy for livestock was significantly worse than for pedestrians. This was likely due to the motion of livestock being dependent both on animal orientation and relationships within a herd, which were not captured in the GRNN model.

Due to the unavailability of a livestock motion model, all simulated trials make use of ORCA pedestrian motion model \cite{VanDenBerg2011} for dynamic agents. Similarly, pedestrians have been used in all real-world trials due to the availability of a response prediction model of person-robot interactions from our prior work \cite{Eiffert2020a} and to allow comparison to simulated results. Additionally, the use of pedestrians in all trials, as opposed to livestock, allows better comparison of our framework to prior state of the art path planners discussed in Section \ref{section:planning}.
Finally, safety of agents is a concern when testing planning approaches in real world trials. The Swagbot robotic platform is a large vehicle weighing over 200 kg and can cause serious bodily injury when travelling at speed, necessitating the use of willing participants until safety around moving individuals can be proven.


\section{Results}
\label{section:results}
Combined results from the simulated testing, extended navigation trial and crowd interaction trial allow a comprehensive evaluation of the performance of our proposed framework in a variety of scenarios. This allows us to evaluate performance in regards to both the ability to safely and effectively navigate through dynamic environments in the presence of moving individuals, and the ability to efficiently follow a reference path during resource constrained navigation.

\begin{figure}[t]
    \centering
	\includegraphics[width=12.5cm,height=7.0cm]{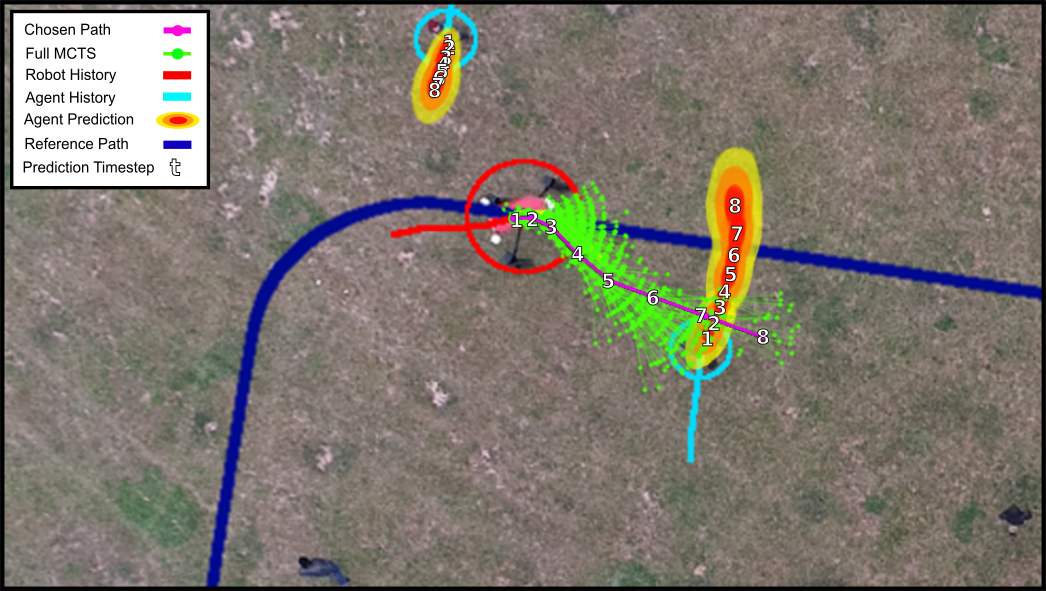}
	\setlength{\belowcaptionskip}{0pt}
	\caption{\textit{An example planning step of the \ac{MCTS}-\ac{GRNN} dynamic planner during the crowd interaction trial, illustrating the full search tree in green and the chosen best path in purple as the robot attempts to navigate to the next supplied position along the reference path. The predicted bivariate Gaussian response of each pedestrian considered during planning is shown as a heatmap extending over 2 standard deviations, based on the observed prior positions. The chosen path and predictions are both shown across 8 timesteps. }}
	\label{tree_search_step}
\end{figure}

\subsection{Dynamic Planning and Collision Avoidance}
\label{collisionresults}
Fig. \ref{tree_search_step} illustrates a single planning step of the navigation framework during the crowd interaction trial, in which the robot is navigating towards the current local goal along the reference path output by the long-term planner, whilst accounting for the nearby observed agents. 
This example shows the chosen path of the robot in purple and the predicted responses of each considered agent as a bivariate Gaussian heatmap over each future timestep. 
The tree search, shown in green, illustrates the exploration of the robot's action space and its consideration of actions that better follow the reference path. As expected, the local dynamic planner takes into account the predicted future motion of the individual moving into the robot's path, choosing a path that will drive the robot behind the predicted travel of the individual to better reach the next local goal in fewer planning steps.

\begin{figure}[H]
    \centering
	\includegraphics[width=16.0cm,height=4.0cm]{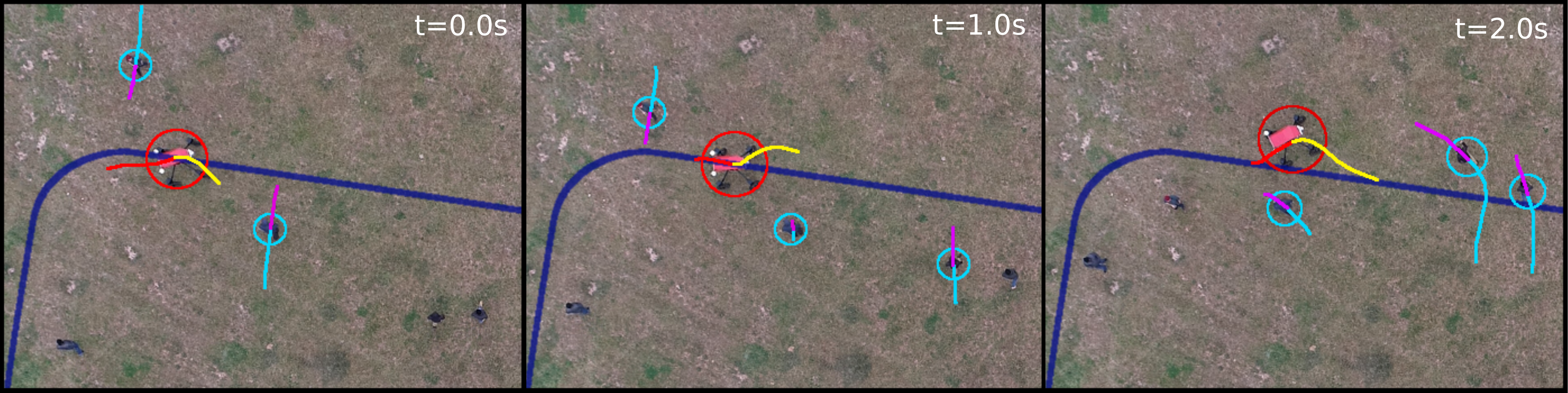}
	\setlength{\belowcaptionskip}{-0pt}
	\caption{\textit{Subsequent time steps to Fig. \ref{tree_search_step} during the crowd interaction trial. The robot's best path found using the \ac{MCTS}-\ac{GRNN} dynamic planner is shown in yellow as it tracks the reference path (dark blue). The mean predicted path of each agent in response to this chosen path is shown in purple. Step 1 shows the robot choosing a path behind the approaching agent. In step 2, when the agent stops rather than following the predicted motion, the robot updates its plan accordingly. In step 3, the robot returns to the reference path, avoiding the new agents. Only agents tracked by the robot each timestep are highlighted. }}
	\label{multiple_steps}
\end{figure}

Fig. \ref{multiple_steps} expands on this example, showing 2 subsequent timesteps of the same interaction. For clarity, these examples are shown at 1 second intervals, rather than the actual planning timestep of 0.2 seconds. The initial step (left) shows the same path as Fig. \ref{tree_search_step}, restricted to 5 planning steps. However, by the next timestep (middle) the motion of the individual ahead of the robot has slowed, resulting in the local dynamic planner updating its proposed plan to travel in front of the individual. The third timestep (right) shows the robot rejoining the reference path once it has effectively navigated through the interaction, again avoiding the newly approaching individuals in the right of the frame.

 \begin{figure}[H]
    \centering
	\includegraphics[width=14.0cm,height=13.0cm]{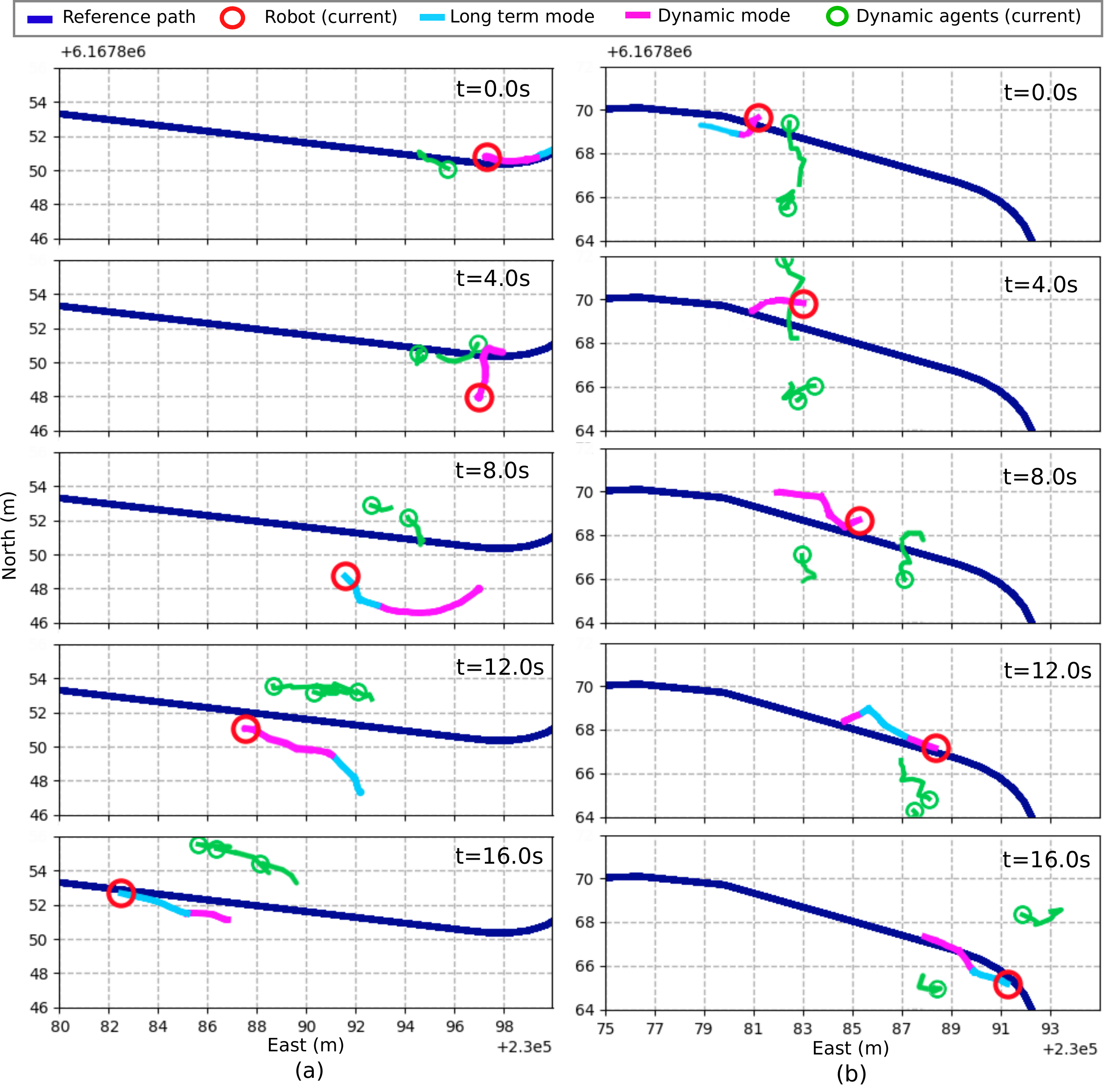}
	\setlength{\belowcaptionskip}{-10pt}
	\caption{\textit{Example scenarios from the crowd interactions trial, illustrating the behaviour of the robot around dynamic agents. Both examples (a) and (b) demonstrate collision avoidance ability of the system when using the MCTS-GRNN local dynamic planner module, as well as the preference to return to the resource efficient long term plan when clear of any agents. Please refer to the supplementary video for more examples.}}
	\label{example_inteactions}
\end{figure}

Further examples of robot crowd interaction are illustrated in Fig. \ref{example_inteactions}, with each column showing a single example over a longer time period in 4 second timesteps. The robot's history is shown in either purple or blue depending on whether it was using the dynamic or long term mode respectively at that time. These examples further demonstrate robot behaviour in the presence of moving individuals, demonstrating how the interplay between each planning mode allows for both efficient navigation through crowds and the return to the reference path when possible. These examples also demonstrate the latching behaviour of the hierarchical mode switcher described in Section \ref{section:highlevelcontrol}. This ensures that the dynamic path is used for a period of time after clearing an interaction to avoid oscillating back and forth between planning modes due to noisy perception of an agent on the edge of the dynamic planning area.

\begin{figure}[t]
    \centering
	\includegraphics[width=15.0cm,height=7.5cm]{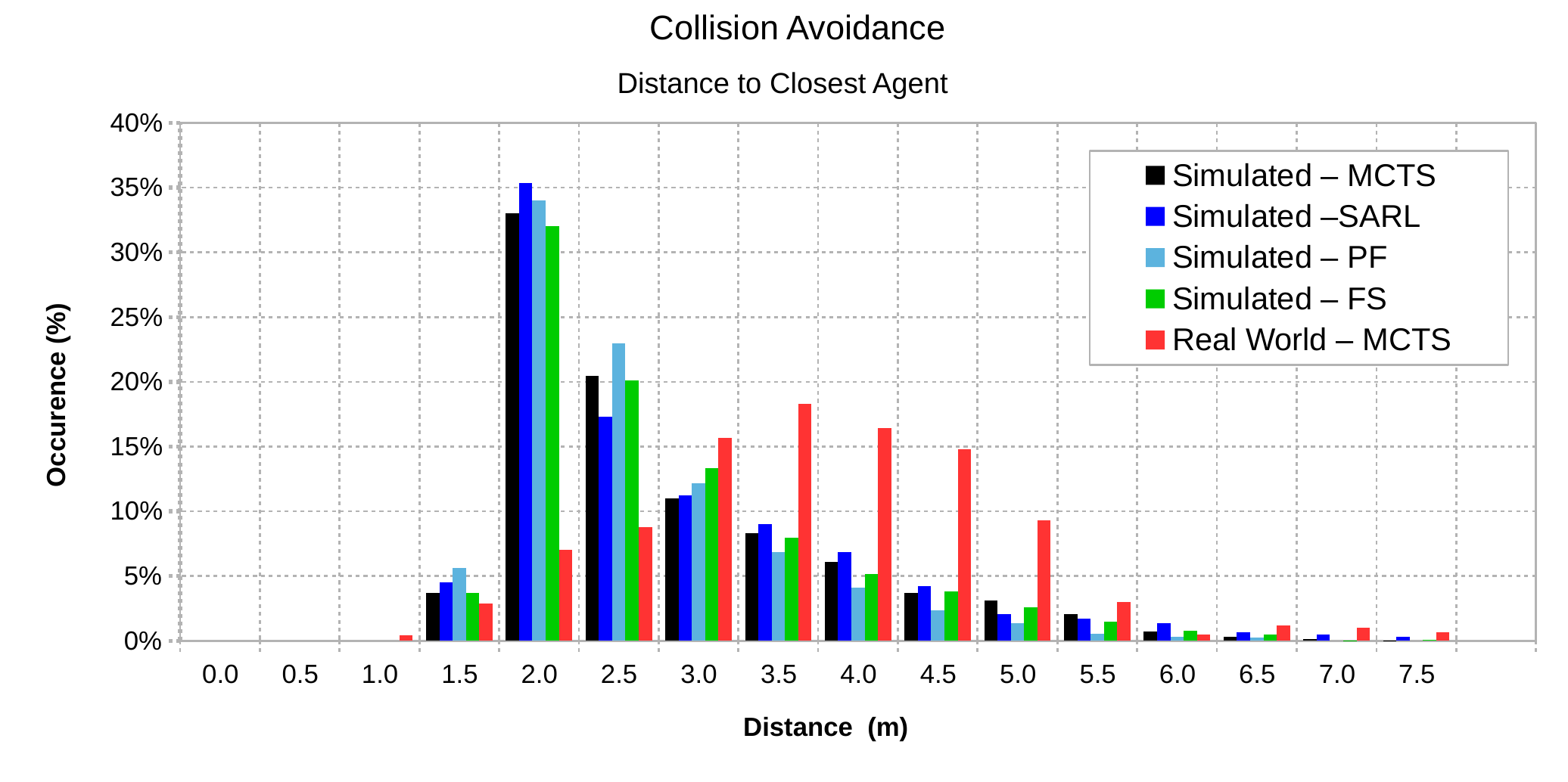}\vspace{0.0cm}
	\setlength{\belowcaptionskip}{-0pt}
	\caption{\textit{A comparison  of the distance to the closest agent throughout all trials, simulated and real, illustrates the ability of each implementation to effectively avoid collisions with moving agents. Distances are show as a histogram of occurrence percentage for all times an agent was within 8 m. Combined results for the extended navigation trial and crowd interaction real-world trials are shown in red. Whilst all simulated planner implementations demonstrate similar ability to avoid collision with agents, there is a significant difference between the real world and simulated results using the same \ac{MCTS} planner implementation, suggesting differences in agent behaviour in the presence of the robot.}}
	\label{min_dist_result}
\end{figure}
     
A quantitative comparison of the results from the real-world tests and the simulated testing is illustrated in Fig. \ref{min_dist_result}. This histogram illustrates the minimum distance between the robot and all surrounding agents, shown as a percentage of occurrence of each distance for all times in which an agent was detected within an 8 m radius of the robot. This distance excludes an additional 1.5 m radius from the robot's centre, which accounts for the physical extent of the robotic platform. This comparison highlights both the safety of the proposed hierarchical approach whilst also indicating differences between the simulated trials and real-world trials in terms of agent behaviour. Please refer to the supplementary video for more examples of robot behaviour and collision avoidance performance in all tested versions.

The real-world results, shown in red, combine both the extended navigation trial and crowd interaction trial. These results demonstrate that the robot is able to effectively maintain a safe distance from agents, keeping an average minimum distance of 3.48 m to the nearest agent. As described in Section \ref{section:approach}, the \ac{FS} collision avoidance module ensures that all robot motion is stopped whenever an obstacle or agent is detected within 2.0 m of the robot (3.5 m from the robot's centre). A single instance of distance less than 1.5 m occurred during real-world testing, where the robot turned on the spot without realising that a person was in its blind spot, as shown in Fig. \ref{robotic_platform}. This was a result of both the limited sensor \ac{FOV}, and a discrepancy between the holonomic dynamics of the real-world robotic platform and the non-holonomic Ackermann assumptions of the dynamic path planner. The simulated results demonstrate significantly less distance maintained between the robot and dynamic agents during testing for all compared framework versions, with average minimum distances of 3.24 m, 3.17 m, 3.20 m, and 3.25 m for the \ac{MCTS}, \ac{SARL}, \ac{PF}, and \ac{FS} methods respectively. Additionally, notable peaks are observed for all versions at the \ac{FS} limit of 2 m. All tested versions were still able to maintain a safe distance from agents, highlighting the safety of the proposed approach even when using local dynamic planner versions that attempt to navigate around agents, and the response aware MCTS-GRNN version which plans with the expectation that agents will respond to its actions.

A comparison of peaks between the results of real-world and simulated \ac{MCTS} trials indicates a difference in agent behaviour. In Section \ref{section:mcts} we describe the state evaluation function used in all \ac{MCTS} tests, where we use a value of $d=2$ in Eq. \ref{eqn:costfn}. This value means that no cost is applied when the robot approaches an agent up to a distance of 2 m, and was used in both real-world and simulated \ac{MCTS} trials. As all simulated agents used a maximum neighbour distance of 1.5 m (excluding agent or robot radii), as described in Section \ref{section:experiments}, the peaks at 2 m observed in simulated trials matches expectations, as the agents approach the \ac{FS} robot limit. Comparatively, the real pedestrians maintain a greater distance on average. While the results show safe distances were maintained during all tests, it may still be preferable to increase the value of $d$ in future experiments to better reflect the preferred distance of real humans for less intrusive robot behaviour.


\subsection{Resource Efficient Navigation}
\label{resourcenavigationresults}

The primary resources considered in this work include the energy usage and time taken during navigation between waypoints.
In real-world applications---such as deployment on agricultural properties for weeding---additional quantities, such as herbicide, would also need to be managed.
Our results again demonstrate that the hierarchical framework can achieve varying resource efficiencies through the use of different dynamic planning modules, and that agent density greatly influences both energy and time efficiency.

\begin{figure}[t]
    \centering
	\includegraphics[width=10.5cm,height=6.5cm]{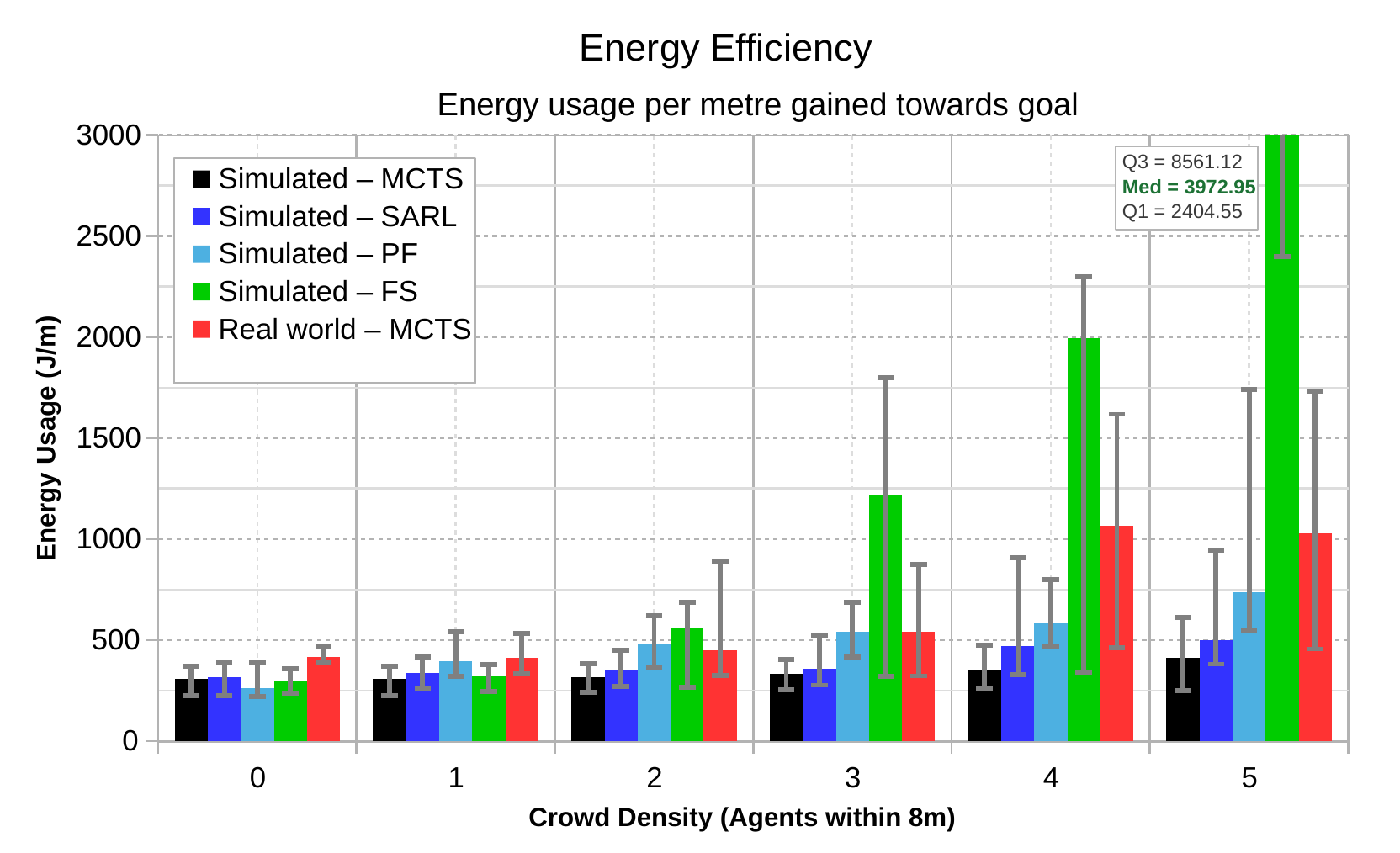}\vspace{0.0cm}
	\setlength{\belowcaptionskip}{-10pt}
	\caption{\textit{Energy efficiency in varying crowd densities, represented by energy used to move 1 m towards the goal (J/m). \ac{MCTS} or \ac{SARL} are able to effectively navigating through denser crowds, using the least energy. The \ac{FS} version uses significantly more energy than any other tested dynamic planning module as crowd density increases. Median values are shown as well as Q1 and Q3 errors.}}
	\label{energy_versus_density}
\end{figure}

Performance is again compared between the real-world tests---the extended navigation and crowd interaction trials---and the simulated trials, with each using either the \ac{MCTS}-\ac{GRNN} dynamic planning module, SARL \cite{chen2019crowd}, the \ac{PF} dynamic planner, or simply the \ac{FS} only.
Fig. \ref{energy_versus_density} illustrates the energy usage of the robotic platform in all trials for increasing crowd densities. This is shown as the energy required to reach a waypoint, normalised for comparison between trials as J/m gained towards goal.

Whilst energy usage is similar between all framework versions when not in the presence of any moving agents, there is significant difference in usage in increasingly dense crowds. The increase in energy usage at a density of 5 agents within 8 m as a percentage of energy usage at a density of zero for \ac{MCTS} and \ac{SARL} versions are just 33.5\% and 58.0\% respectively, whereas \ac{PF} increases 178\%  and \ac{FS} version 1230\%. The immense increase in \ac{FS} is a result of it sitting stationary for extended periods of time waiting for agents to move and the base power draw of 203W for the robotic platform, detailed in Section \ref{section:experiments} . The real world \ac{MCTS} version uses 148\% of it's baseline energy usage, significantly more than the simulated version, suggesting a greater tendency to rapidly change velocity. This may be a result of the real world perception not detecting agents until they are much closer to the robot, as discussed in Section \ref{perceptionresults}.

\begin{figure}[t]
    \centering
	\includegraphics[width=9.5cm,height=6.0cm]{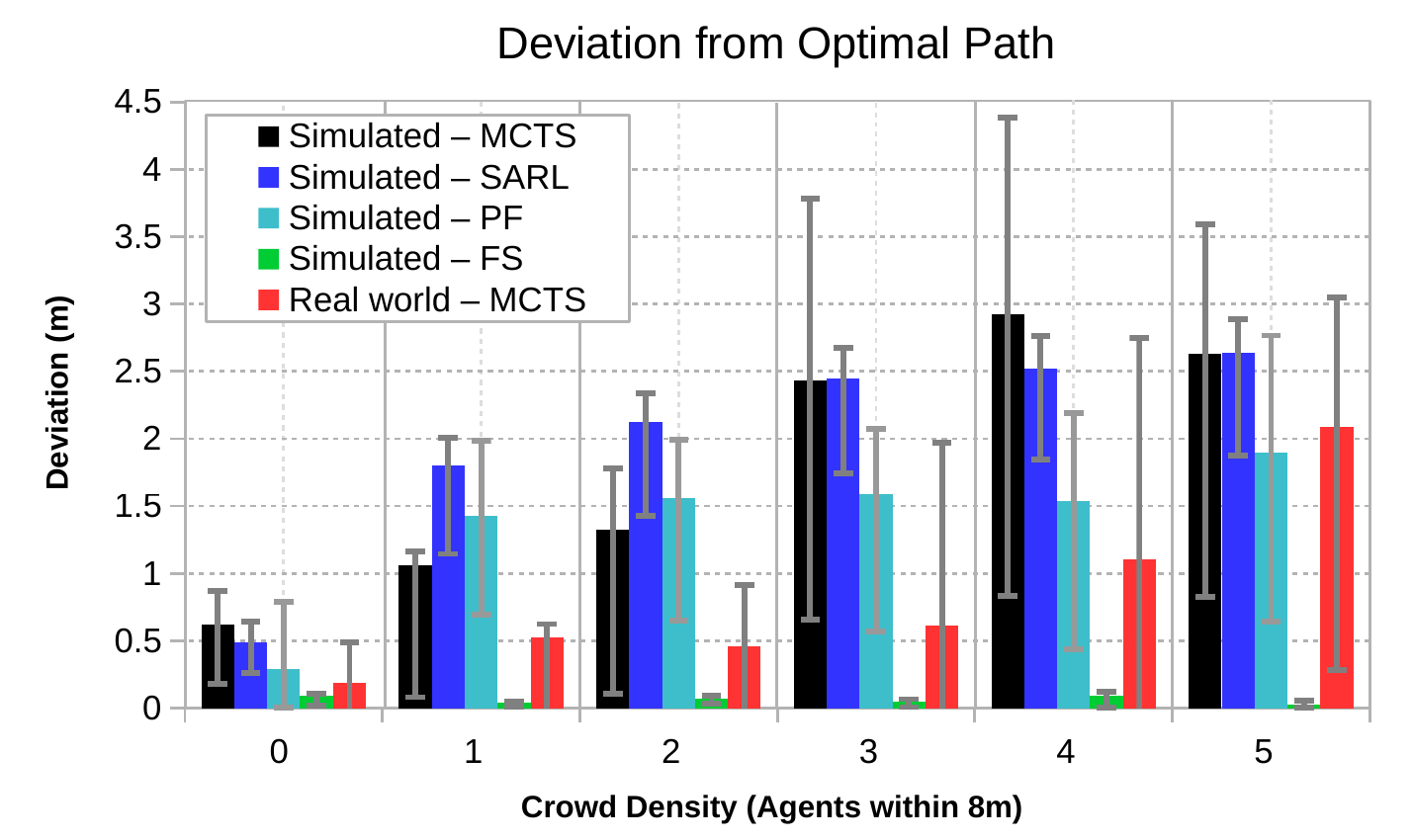}\vspace{0.0cm}
	\setlength{\belowcaptionskip}{-10pt}
	\caption{\textit{Deviation from the optimal energy efficient path in varying crowd densities for each framework implementation. This is an important consideration in environments with large changes in elevation, where deviating from the reference path can lead to significantly large energy use. FS version does not deviate from the reference path at all. PF deviates significantly less than both MCTS and SARL versions. Median values are shown as well as Q1 and Q3 errors.}}
	\label{deviation_versus_density}
\end{figure}

All trials undertaken in this work use environments with less than 5 m maximum elevation change and no slopes with a gradients over 6 $^{\circ}$. However, in environments with significantly steeper slopes, deviation from the optimal reference path can lead to much greater changes in energy use. Fig. \ref{deviation_versus_density} shows the ability of each tested framework implementation to accurately track the reference path in increasing crowd densities. Whilst the \ac{FS} and \ac{PF} versions use more energy in the simulated trials than \ac{MCTS} or \ac{SARL}, they also deviate significantly less in denser crowds, with the \ac{FS} version never deviating more than 0.2 m from the reference path. As the \ac{FS} implementation demonstrates expected minimal deviation, these results are considered a baseline for comparison. This results suggests that in more difficult terrain, the simpler planners may lead to decreased energy usage.

\begin{figure}[H]
    \centering
	\includegraphics[width=10.5cm,height=6.0cm]{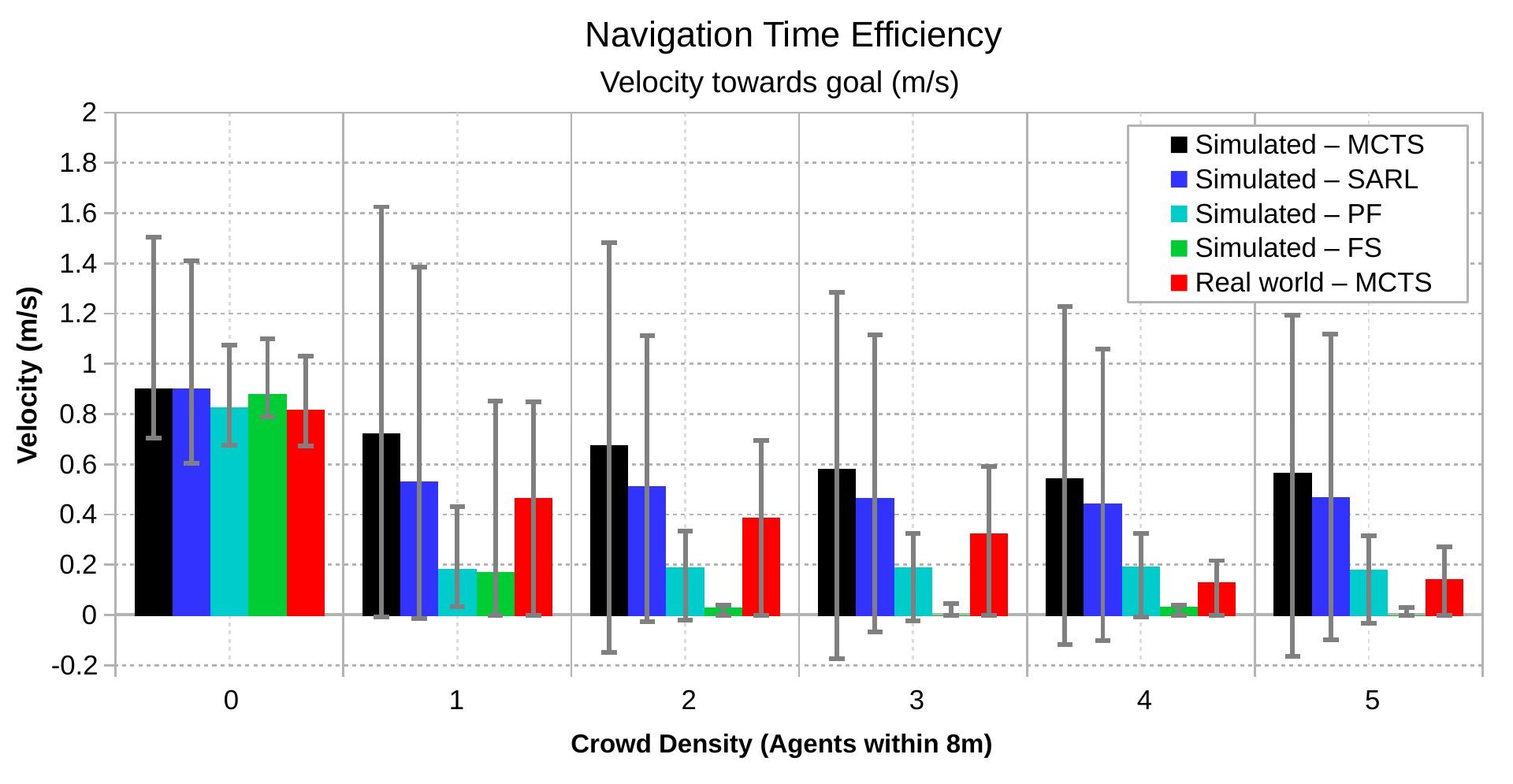}\vspace{0.0cm}
	\setlength{\belowcaptionskip}{-10pt}
	\caption{\textit{Navigation time efficiency in varying crowd densities, represented by velocity towards the goal. Both the crowd response aware \ac{MCTS} and \ac{SARL} implementations of the planning framework are better able to continue navigating towards the goal as the crowd density increases compared to the \ac{FS} and PF versions in simulation. The \ac{FS} version approaches a velocity of 0 m/s in density=2, spending the majority of its time stationary. Median values are shown as well as Q1 and Q3 errors.}}
	\label{speed_versus_density}
\end{figure}

Fig. \ref{speed_versus_density} compares the robot's navigation time efficiency against crowd density, in terms of velocity (m/s) towards the goal. All tested methods achieve a similar velocity towards goal of between 0.8 and 0.9 m/s median value when there are no agents present. However, a large drop in performance of 80.7\%---a reduction in goal approach velocity from 0.88 to 0.17 m/s---is seen in the baseline \ac{FS} version when only a single agent is introduced. A similar drop in performance of 77.8\% is seen with the non response-aware \ac{PF} approach. Comparatively, decreases of just 23.0\% and 38.3\% are seen when using the \ac{MCTS} and \ac{SARL} versions respectively in the same simulated environment. This result is reflected in the real-world trials where a decrease of just 43.0\% is seen for \ac{MCTS}. Additionally, the \ac{FS} version approaches a velocity of 0 m/s with just 2 agents within an 8 m radius of the robot. Whilst the \ac{PF} and \ac{MCTS} versions deviate significantly from the optimal path in the presence of increasing crowd density, they are able to continue making progress towards the goal even in the most dense tested crowd of 5 agents within the robot's vicinity. 

These results suggest that resource efficiency tradeoffs are dependent both on the choice of module used within the framework, and the expected environment. Whilst the use of a local dynamic planner better able to navigate crowds can greatly decrease the total time taken, resulting in significant energy savings, it also has the potential to greatly increase energy usage in uneven terrain.

\subsection{Object Detection and Tracking}
\label{perceptionresults}
\begin{figure}[t]
    \centering
	\includegraphics[width=12.5cm,height=8.8cm]{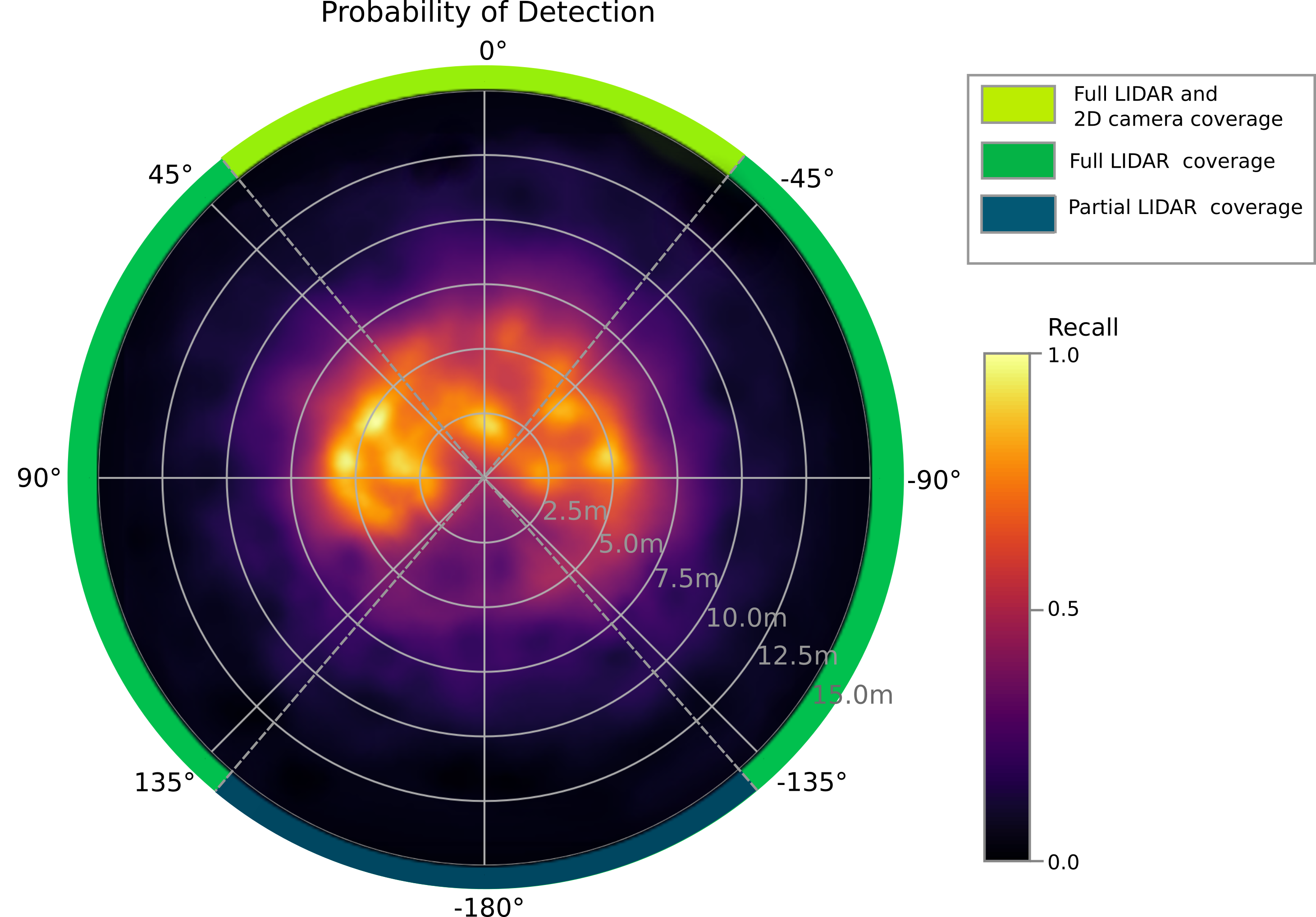}\vspace{0.0cm}
	\setlength{\belowcaptionskip}{10pt}
	\caption{\textit{Probability of detecting a present agent (recall) within a 15 m radius of the robot, based on the output of the perception pipeline evaluation as per Section \ref{section:perception_analysis}. The robot's sensor FOV is shown around the heatmap perimeter, with the 2D camera covering $\pm38^{\circ}$ and the \ac{LIDAR}  partially obstructed beyond $\pm 140^{\circ}$.}}
	\label{detection_comaprison}
\end{figure}

Fig. \ref{detection_comaprison} illustrates ability of the perception pipeline used within the hierarchical framework to correctly detect the location of nearby agents. Recall---representing the probability that a present agent will be detected---is shown up to a range of 15 m across the robot's sensing space.
As shown in Fig. \ref{robotic_platform}, the robot's sensor \ac{FOV} is limited, with the 2D camera covering only part of the forward-facing quadrant, from $-38^{\circ}$ to $+38^{\circ}$ and the \ac{LIDAR}  partially obstructed beyond $\pm 140^{\circ}$ by the robot's frame and legs. Whilst the angular distributions of Fig. \ref{detection_comaprison} match expectations based on the \ac{FOV}, there exist a number of missed detections directly in front of the robot within a range of 5 m. Out of a total of 1406 missed detections, 1 occurred within 2.5 m, and 15 occurred within 5 m in the robot's forward-facing quadrant. This result emphasises the importance of using a \ac{FS} collision avoidance system which is not reliant on synchronised association between \ac{LIDAR} and 2D camera, nor on 2D detections of objects. Instead, it directly uses the output of the \ac{LIDAR} after removal of the ground plane, as shown in Fig. \ref{perception_pipeline}.

Table \ref{tab:perceptionresults} summarises both precision and recall across the sensing space. Whilst recall gives an indication of the safety of the system, precision provides a measure of how often false positives occur --- important when considering the efficiency of the overall system. As shown in Section \ref{resourcenavigationresults}, efficiency in terms of both time and energy usage decrease in the presence of more dynamic agents for all tested planner versions. As such a minimisation of false detections will lead to more efficient resource usage as the robot does not have to react to non-existent obstacles. The precision of the tested perception pipeline is high up to 8 m in both the front quadrant, where 2D and \ac{LIDAR} sensing is available, and the side quadrants, with only \ac{LIDAR}, but drops significantly in the rear quadrant. This suggests that the decision to limit consideration of detections to just the forward and side facing areas described in Section \ref{section:highlevelcontrol} for both the fail safe module and dynamic planners was beneficial to resource efficiency.




\begin{table}[]
\centering
\begin{tabular}{l|l|l|l|l|l|l|}
\cline{2-7}
\textbf{}                            & \multicolumn{3}{l|}{\textit{\textbf{Recall}}} & \multicolumn{3}{l|}{\textit{\textbf{Precision}}} \\ \cline{2-7} 
\textit{Quadrant}                            & \textbf{0-4 m}  & \textbf{4-8 m} & \textbf{8 m+} & \textbf{0-4 m}   & \textbf{4-8 m}  & \textbf{8 m+}  \\ \hline
\multicolumn{1}{|l|}{\textbf{Front}} & 0.850          & 0.720         & 0.057        & 1.00           & 1.00          & 0.77         \\ \hline
\multicolumn{1}{|l|}{\textbf{Sides}}  & 0.642          & 0.510         & 0.033        & 0.974           & 0.982          & 0.72         \\ \hline
\multicolumn{1}{|l|}{\textbf{Rear}}  & 0.097          & 0.168         & 0.00        & 0.031           & 0.424          & N/A            \\ \hline
\end{tabular}
\caption{\textit{Recall and precision of the perception pipeline across the robot's sensing space, based on distance (0-4 m, 4-8 m and 8 m+) and quadrant (front: $|\phi| \leq 45^\circ$; sides: $135^\circ \geq |\phi| >45^\circ$; rear: $|\phi| > 135^\circ$, where $\phi$ is as per Fig. \ref{detection_comaprison}). Note that no detections exist in the rear quadrant at distances beyond 8 m.}}
\label{tab:perceptionresults}
\end{table}

\newpage

Additionally, Fig. \ref{detection_comaprison} highlights the limitations of the robot's ability to accurately observe the state of any crowd or herd it is within. This limitation should be considered both during planning around moving individuals and during the training of any predictive model of agent motion that will use the robot's current observed state during real-world inference.
A better understanding of agent observation probability across the robot's planning space could be used to inform a robot when planning in crowded environments.  In both simulated trials and real-world trials carried out during this work, the robot's motion was restricted to the forward quadrant due to sensor \ac{FOV}. However, Fig. \ref{detection_comaprison} demonstrates relatively high observation likelihood extending beyond this quadrant into planning space covered only by \ac{LIDAR}, suggesting that movement in these directions should also be allowed by the local dynamic planning module.
 
Whilst predictive models such as that used by the \ac{MCTS}-\ac{GRNN} dynamic planner are trained using full knowledge of nearby individuals, this is not the case in real-world implementations in which the input to these models is limited to only the robot's observations. These observations are limited by the FOV of the robot's on board sensors and possible occlusions in a crowd, as shown in Fig. \ref{robotic_platform}. As these models are intended for use in predicting the response of an agent to a robot's motion, they will invariably be incorrect when the agent is reacting to other unobserved agents. Similarly, the models will be incorrect when unable to observe the complete history of an agent due to missed detections. By instead training these models using only the position of agents observable to the robot as input, they will better reflect real-world use. As the ground truth motion of each agent will still reflect its response to other unobserved agents, these models may better learn to predict the motion of agents in partially observed crowds or herds and assign greater uncertainty in situations without full observability of a crowd, as will be experienced in real robotic implementations.

\section{Discussion}
\label{section:discussion}

\subsection{Offline Crowd Density Consideration}
\label{section:discussionoffline}
The results in Section \ref{resourcenavigationresults} highlight the influence of nearby agent density on both the energy and time efficiency during navigation. This suggests that the expected crowd or herd density that the robot will be operating within should be a consideration during the initial offline planning step. 
A priori knowledge of areas in the environment where energy efficient motion may be compromised due to the presence of crowds---which often necessitate deviation from the optimal path to navigate safely---would allow for better estimates of the energy expenditure for a given plan. Over extended missions this will better allow the robot to determine when it needs to return to charging stations, and should also be a consideration alongside terrain in the energy cost of motion model used to compute the optimal path. Alternatively, it would allow for the inclusion of a factor of safety in resource usage when operating in areas with unknown possible crowd densities.

\subsection{Online Adaptive Framework}
\label{section:adaptive}
In all trials undertaken in this work the hierarchical framework was deployed using a single type of local dynamic planner module in each implementation. However, the hierarchical mode switcher is currently structured to take input from multiple planners simultaneously.
By allowing the type of dynamic planner used to vary during operation, the framework could allow changing behaviour, optimising for different resource constraints as required. 
A common example would be when it is desirable for a robot to use its resources efficiently while also meeting a deadline. While the simpler \ac{FS} and \ac{PF} planners would reduce energy costs from path deviation in the case of undulating terrain, this comes at the cost of time, and in many circumstances---such as in crowded environments---this trade-off can quickly become unfavourable.
By switching to a dynamic planner better able to navigate crowds, such as the \ac{MCTS}-\ac{GRNN} or \ac{SARL} versions, the behaviour of the robot could be easily changed to take a faster path away from the energy optimal path, at the possible expense of energy due to variations in terrain.


\subsection{Predictive Model Limitations}
\label{section:limitations}
The predictive model used within the tested \ac{MCTS}-\ac{GRNN} local dynamic planner was trained on a dataset of robot-pedestrian interactions obtained in a semi-structured shared road environment. Additionally, it was clear to the neighbouring pedestrians that the vehicles used in the dataset were all human-operated. 
Whilst the pedestrians in this work were instructed to treat the mobile robot as though it were also human-operated, it is unlikely that their behaviours and responses to the robot's motion would have remained the same as the pedestrians in the training dataset, leading to inaccuracies in the predictions of the local dynamic planner. Due to different social cues and norms between groups of humans and animals it is likely that any predictive model will only learn the response of the training population to the observed robot type and behaviour, and would experience distribution shift even as the training group became acquainted with the robot throughout testing. To overcome this, an online version of the predictive model would be required, which can update based on observed differences between the predicted and actual motion of nearby agents during interactions.
\section{Conclusion}
\label{section:conclusion}

Through the analysis of simulated and real world trials---both prior and new---this work has demonstrated how a resource limited mobile robot can achieve extended autonomy for large scale farming. This has been achieved through the use of our proposed planning framework, based on the sum of prior work, which allows both resource aware and crowd response aware path planning in real world unstructured and dynamic environments.


The trials undertaken have involved both the weeding of pastures alongside moving individuals and navigation through more densely populated environments. Analysis of results, based on resource usages of energy and time and the ability to effectively navigate crowds, have shown efficient resource usage as well as its safety around moving individuals. Comparisons of the framework when using varied local dynamic planning methods in simulation have also demonstrated how resource usage can be adapted to suit the environment without adversely impacting safety. Whilst the response aware \ac{MCTS}-\ac{GRNN}  and state of the art reinforcement learning approach \ac{SARL} version of the framework allowed for improved navigation time efficiency, they deviated significantly more from the optimal energy efficient path than the simpler \ac{PF} and baseline \ac{FS} versions. Whilst this did not lead to increased energy usage in the relatively uniform terrain of this work, in more uneven terrain this could lead to significantly greater energy usage and should be a consideration when choosing the appropriate planning strategy.

Finally, an evaluation of the tested perception pipeline used in all real-world trials has provided an understanding of detection likelihood around the mobile robot. This result has suggested how an understanding of agent observation likelihood across the robot's planning space could be used to improve planning in crowded environments both by allowing for the learning of improved models of agent motion in partially observable crowds, and by better informing sampling-based planners during a search of the agent's action space.

Avenues for future work include the integration of recharging \cite{Wallace_CASE2020} into the framework, as well as direct consideration of energy costs in the local planning stage, both offline as discussed in Section \ref{section:discussionoffline}, and online as discussed in Section \ref{section:adaptive}. Additionally, adding a subsequent trajectory refinement step to the hierarchical planning framework, leveraging strategies such as those presented in \cite{Gao2020}, may help further improve path smoothness and motion efficiency. Finally, improvement to the perception pipeline to allow estimation of livestock orientation, as well as improved motion prediction models which can capture intra-herd dependencies through attention mechanisms plan to be implemented in order to allow livestock response prediction in the presence of a mobile robot.

\section*{Appendix}
A multimedia file is included as an appendix to this work. This file comprises a video containing an overview of all experiments presented in this work, and additional qualitative examples of robot behaviour. The video is also available at https://youtu.be/DGVTrYwJ304. 
\section*{Acknowledgements}
The authors would like to thank everybody from the ACFR involved in the development of the Swagbot Robotic Platform, those who helped during the field trials conducted during this work---specifically Khalid Rafique, Javier Martinez, Thomas Ingram, and Jeremy Randle---and to Asher Bender for his valuable advice.


\bibliographystyle{apalike}

\bibliography{jfrRefs}

\end{document}